\documentclass[10pt,twocolumn,letterpaper]{article}

\usepackage{cvpr}              %

\definecolor{cvprblue}{rgb}{0.21,0.49,0.74}
\usepackage[pagebackref,breaklinks,colorlinks,allcolors=cvprblue]{hyperref}
\usepackage[utf8]{inputenc} %
\usepackage[T1]{fontenc}    %
\usepackage{hyperref}       %
\usepackage{url}            %
\usepackage{booktabs}       %
\usepackage{amsfonts}       %
\usepackage{nicefrac}       %
\usepackage{microtype}      %
\usepackage{xcolor}         %
\usepackage{colortbl}
\usepackage{graphicx}
\usepackage{multicol}
\usepackage{multirow}
\usepackage{makecell}
\usepackage{tablefootnote}
\usepackage{amsmath}
\usepackage{amssymb}
\usepackage{tikz}
\usepackage{xspace}
\usepackage{duckuments}    %
\usepackage{wrapfig}
\usepackage{caption}
\usepackage{algorithm}
\usepackage{dsfont}
\usepackage{algpseudocode}
\usepackage{algorithmicx}
\usepackage{bbm}

\usepackage{tcolorbox}
\usepackage{enumitem}
\usepackage{color}
\usepackage{lipsum} %
\usepackage{pifont} %

\usepackage{pgfplots}
\pgfplotsset{compat=1.18}

\input{preambles/macro}

\def\paperID{3090} %
\def\confName{CVPR}
\def\confYear{2025}

\title{\nickname: Foundation Dataset and Model for Open-Vocabulary 3D Segmentation}

\author{
Junha Lee$^{1,2,\dagger}$ \qquad
Chunghyun Park$^{1,2,\dagger}$ \qquad
Jaesung Choe$^{1}$ \\
Yu-Chiang Frank Wang$^{1}$ \qquad
Jan Kautz$^{1}$ \qquad
Minsu Cho$^{2}$ \qquad
Chris Choy$^{1}$ \\
\\
{$^{1}$NVIDIA} \quad
{$^{2}$POSTECH}
}

\begin{document}
\twocolumn[{%
    \maketitle
    \renewcommand\twocolumn[1][]{#1}%
    \vspace{-5mm} 
    \centering
    \includegraphics[width=.999\linewidth]{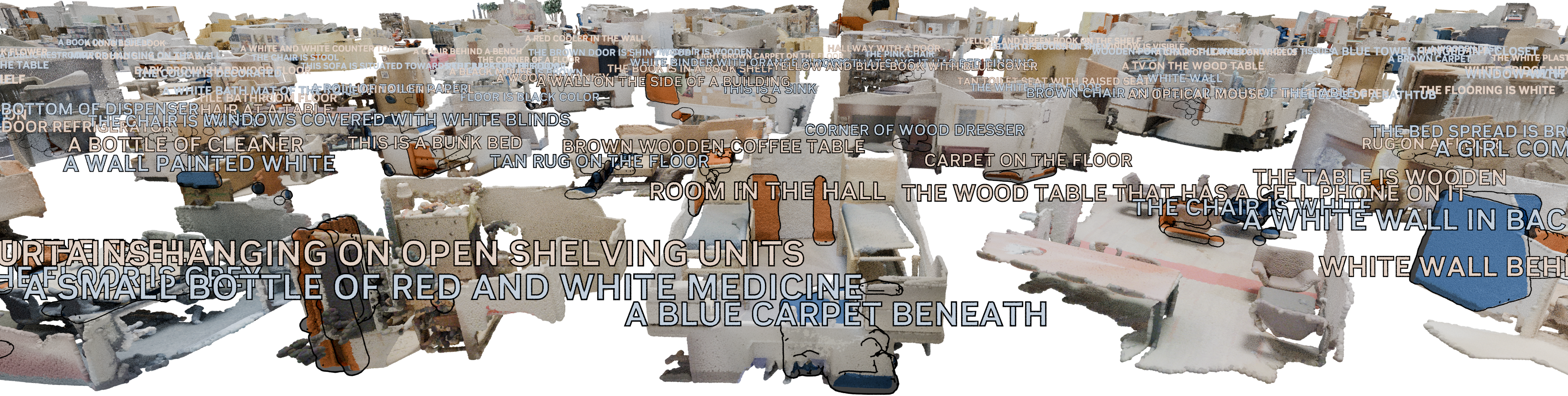}
    \captionof{figure}{
        \textbf{\dataname.}
        \dataname is a large-scale dataset generated from a collection of existing datasets~\cite{dai2017scannet,baruch2021arkitscenes,yeshwanth2023scannet++,chang2017matterport3d,zheng2020structured3d}, consisting of 5.6M mask-text pairs, providing fine-grained masks (black outline in the figure) and detailed captions (text with matching color) pairs.
        Using this large-scale dataset, we propose \nickname, a foundation model for open-vocabulary 3D segmentation.
        \label{fig:teaser}
    }
    \vspace{4mm}
}]

\begin{abstract}

We tackle open-vocabulary 3D scene segmentation tasks by introducing a novel data generation pipeline and training framework.
Our work targets three essential aspects required for an effective dataset: precise 3D region segmentation, comprehensive textual descriptions, and sufficient dataset scale.
By leveraging state-of-the-art open-vocabulary image segmentation models and region-aware vision-language models (VLM), we develop an automatic pipeline capable of producing high-quality 3D mask-text pairs.
Applying this pipeline to multiple 3D scene datasets, we create \dataname, a dataset of more than 30K annotated scenes with 5.6M mask-text pairs - significantly larger than existing datasets.
Building on these data, we propose \nickname, a 3D visiual foundation model (3D-VFM) combining a 3D encoder trained with contrastive learning and a lightweight mask decoder for open-vocabulary 3D semantic and instance segmentation.
Our approach achieves state-of-the-art results on open-vocabulary 3D semantic and instance segmentation benchmarks including ScanNet200, Matterport3D, and ScanNet++, with ablation studies validating the effectiveness of our large-scale training data.
\url{https://nvlabs.github.io/Mosaic3D/}
\blfootnote{
$^{\dagger}$Authors equally contributed to this work during internship at NVIDIA.
}

\end{abstract}

\begin{figure}[t]
    \centering
    \includegraphics[width=\linewidth]{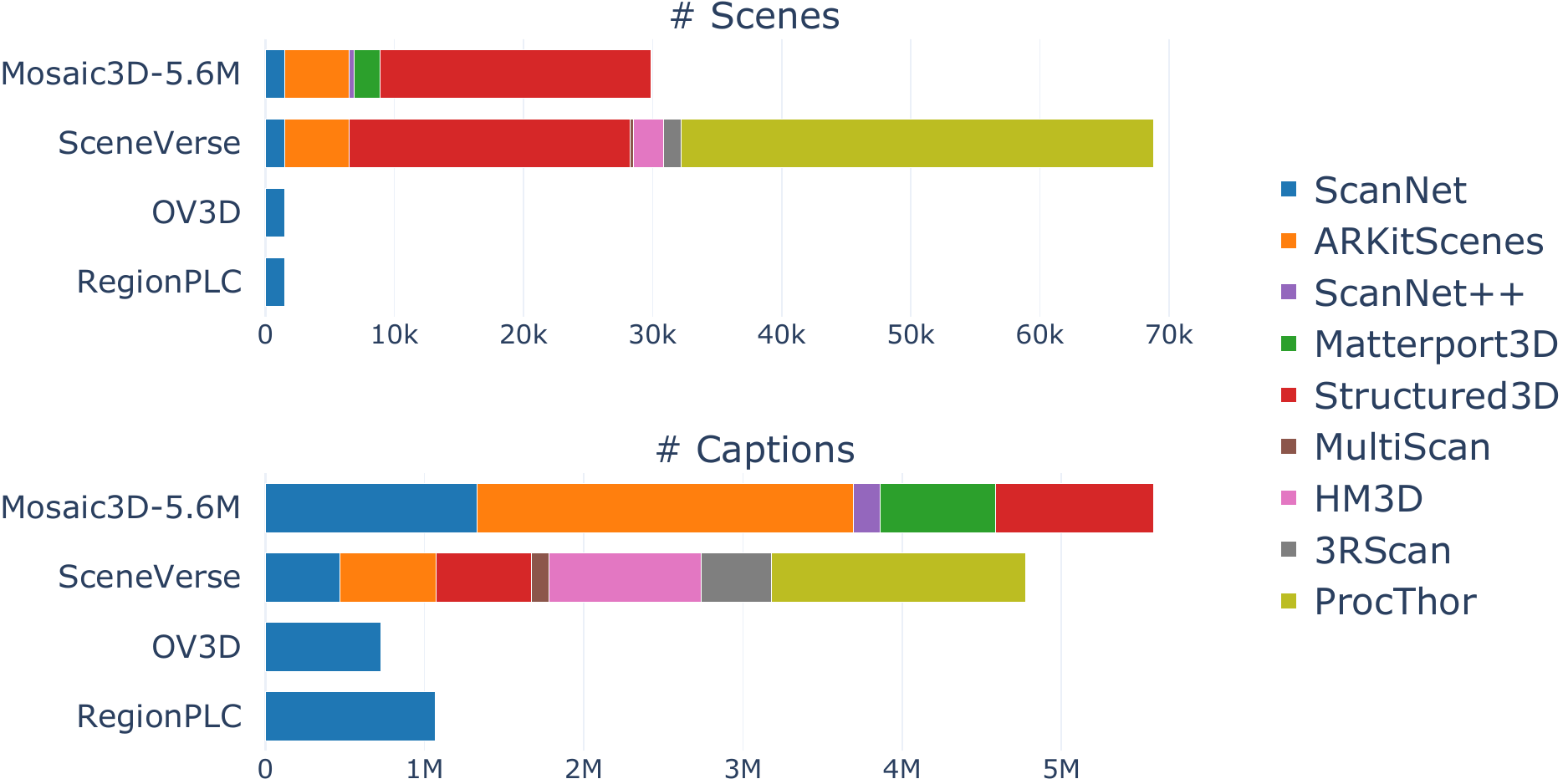}
    \vspace{-5mm}
    \caption{\textbf{Statistics of 3D mask-text datasets.} We show the total number of scenes, tokens for generated captions. Our \dataname significantly surpasses previous datasets in scale, combining multiple datasets to create the largest 3D mask-text dataset to date.}
    \label{fig:data_stats}
\end{figure}

\notoc{\section{Introduction}
\label{sec:intro}

3D scene understanding is a fundamental problem in computer vision that involves detecting and localizing objects while comprehending complex spatial relationships in 3D environments.
This capability is essential for various applications, including robotics, AR/VR, human-computer interactions, and autonomous vehicles.
While traditional approaches rely on predefined object categories, the field is evolving toward open-vocabulary 3D scene understanding, where systems can recognize arbitrary concepts without being constrained to the predefined label sets.
Despite humans' innate ability to perform such tasks effortlessly, developing comparable machine capabilities remains an open problem.

The key bottleneck in advancing open-vocabulary 3D scene understanding is the scarcity of large-scale, high-quality training data.
This limitation is particularly striking compared to 2D vision-language models~\cite{radfordLearningTransferableVisual2021,hu2022scaling,li2022language,ghiasi2022scaling,zhai2023sigmoid,li2024if,liBLIPBootstrappingLanguageImage2022,liBLIP2BootstrappingLanguageImage2023,guopen,liang2023open,rao2022denseclip,shafiullahclip,xu2022groupvit}, 
which have achieved remarkable open-vocabulary capabilities through training on web-scale image-text pair datasets~\cite{radfordLearningTransferableVisual2021,schuhmann2022laion,kakaobrain2022coyo-700m,chenpali,gadre2024datacomp}. 
Unfortunately, creating datasets of comparable scale for 3D scenes remains prohibitively expensive and time-consuming.

Training effective open-vocabulary 3D scene understanding models requires datasets that satisfy three critical requirements:
(1) precise 3D region annotations that delineate object boundaries,
(2) rich textual captions that characterize the visual and semantic attributes of each region, and
(3) substantial scale to encompass diverse domains and offer rich visual-semantic variations.
Creating such data manually, however, becomes increasingly intractable as datasets grow.

To address these challenges, recent works~\cite{ding2022pla,yang2024regionplc,jiang2024open} leverage 2D visual foundation models (VFM)~\cite{llava,vit-gpt2,peng2023kosmos,zou2024segment} to automate data annotation.
They generate 3D mask-text pairs on multi-view RGB-D frames using 2D VFMs and aggregate the generated captions in 3D space.
However, they fall short in meeting aforementioned requirements: 
they use coarse bounding box detectors~\cite{ding2022pla,yang2024regionplc}; 
only have simple attribute labels~\cite{jiang2024open}; and are limited in scale, containing only a few thousand scenes, as shown in Fig.~\ref{fig:data_stats}.
Existing 3D-text pair datasets for 3D vision-language models~\cite{jiaSceneVerseScaling3D2024,wangEmbodiedScanHolisticMultiModal2023,lyu2024mmscan} also face limitations in both the richness of textual captions and the precision of 3D masks, primarily because they rely on human-annotated object labels and 3D region annotations.

In this paper, we address these limitations by introducing an improved data generation pipeline to create high-quality, large-scale 3D mask-text pairs.
Our pipeline satisfies all three criteria by leveraging state-of-the-art open-vocabulary image segmentation models~\cite{liu2023grounding,sam,ravi2024sam,zou2024segment} for precise region segmentation and advanced region-aware vision-language models~\cite{yuan2024osprey} for generating comprehensive textual captions at scale.
By applying this pipeline to a diverse collection of 3D scene datasets~\cite{dai2017scannet,yeshwanth2023scannet++,baruch2021arkitscenes,chang2017matterport3d,zheng2020structured3d}, we create \textbf{\dataname}, a large dataset containing over 30K scenes with 5.6M region captions -- significantly exceeding existing datasets~\cite{yang2024regionplc,jiang2024open} in scale -- while maintaining high-quality region masks and detailed textual captions.

Building upon this new dataset, we analyze how improving the annotation quality and scaling up the data impact open-vocabulary 3D scene segmentation.
To enable this analysis, we develop a general framework for open-vocabulary 3D semantic and instance segmentation.
We train our foundational 3D encoder, \textbf{\nickname}, which aligns a per-point feature with a text embedding vector through contrastive learning.
Then, we train a lightweight mask decoder to predict object instances directly from language-aligned features, enabling the first single-stage open-vocabulary 3D instance segmentation without ground truth labels.
Our approach achieves state-of-the-art results on multiple semantic and instance segmentation benchmarks.
Extensive ablation studies show that both the scale and the quality of our dataset are crucial factors that contribute to the superior performance of our approach.
}
\begin{figure*}[t!]
    \centering
    \includegraphics[width=\linewidth]{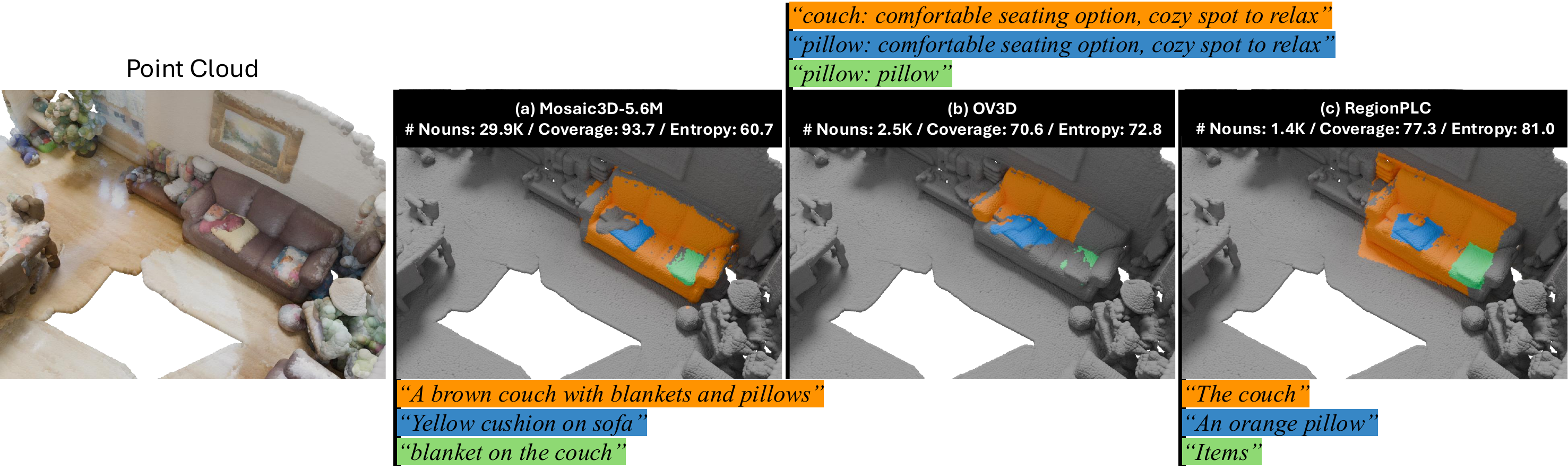}
    \vspace{-7mm}
    \caption{
        \textbf{Dataset comparison.}
        We compare datasets using three metrics: \textit{\# Nouns} (the total number of unique normalized nouns in captions; higher is better), \textit{Coverage} (the percentage of 3D points with associated captions per scene; higher is better), and \textit{Entropy} (the entropy of GT instance ID distribution within masks; lower means more homogeneity - hense better).
        \textbf{(a)} \dataname uses precise masks (Entropy: 60.7) with region-aware VLMs for detailed descriptions (\# Nouns: 29.9K).
        \textbf{(b)} OV3D~\cite{jiang2024open} produces simple attribute labels (\# Nouns: 2.5K) lacking comprehensive visual descriptions.
        \textbf{(c)} RegionPLC~\cite{yang2024regionplc} uses coarse bounding boxes, yielding imprecise masks (Entropy: 81.0).
    }
    \label{fig:data_comparison}
\end{figure*}

\notoc{\section{Related Work}
\label{sec:related}

\noindentbold{2D visual foundation models}
In recent years, we have witnessed the emergence of large pretrained models—so-called foundation models that are trained on large-scale datasets and serve as a \textit{foundation} for many downstream tasks.
These models demonstrate remarkable versatility across multiple modalities, including language~\cite{team2023gemini,touvron2023llama,touvron2023llama2,dubey2024llama3,vicuna2023,radford2019language,brown2020language,chung2024scaling,achiam2023gpt,bai2023qwen,yang2024qwen2,jiang2023mistral,jiang2024mixtral}, vision~\cite{sam,ravi2024sam,dino_v1,oquab2023dinov2,zou2024segment,rombach2022high,ho2020denoising,nichol2021improved,songdenoising,songscore}, audio~\cite{deshmukh2023pengi,zhang2023speechgpt,rubenstein2023audiopalm,borsos2023audiolm}. 
Furthermore, they enable multi-modal reasoning capabilities that bridge across different modalities~\cite{girdharImageBindOneEmbedding2023,Qwen-VL,llava,radfordLearningTransferableVisual2021,jia2021scaling,team2024gemini}.
Among these models, those that operate on visual modalities are known as visual foundation models (VFM).
VFMs excel in various computer vision tasks such as image segmentation~\cite{sam,ravi2024sam,zou2024segment,zou2023generalized,cheng2021per,cheng2022masked,jain2023oneformer,li2024semantic}, object detection~\cite{liu2023grounding,carion2020end}, representation learning~\cite{dino_v1,oquab2023dinov2}, and open-vocabulary understanding~\cite{radfordLearningTransferableVisual2021,li2022language,ghiasi2022scaling,ram,ram_pp,yu2023convolutions,kang2024defense,naeem2024silc,cho2024cat}.
When integrated with large language models, they enable sophisticated visual reasoning and natural language interactions~\cite{llava,Qwen-VL,girdharImageBindOneEmbedding2023,team2024gemini,guo2024regiongpt,yuan2024osprey,you2023ferret}.
We use such vision language models to construct open vocabulary segmentation and captions for point clouds based on multiview images.

\noindentbold{Open-vocabulary 3D segmentation}
Building on the success of 2D VFMs, recent work have extended open-vocabulary capabilities to 3D scene understanding.
OpenScene~\cite{Peng2023OpenScene} first introduced zero-shot 3D semantic segmentation by distilling knowledge from language-aligned image encoders~\cite{li2022language,ghiasi2022scaling}.
Subsequent methods~\cite{ding2022pla,yang2024regionplc,jiang2024open} leverage multiview images to generate textual captions, which then serve as training supervision.
However, these methods face challenges in generating high-quality 3D mask-text pairs at scale.
For open-vocabulary 3D instance segmentation, existing methods~\cite{takmaz2023openmask3d,nguyen2024open3dis,huang2024openins3d} typically rely on closed-vocabulary proposal networks such as Mask3D~\cite{schult2023mask3d}, which inherently constrains their ability to detect novel object categories. 
Moreover, these methods leverage 2D VFMs like CLIP~\cite{radfordLearningTransferableVisual2021} for region classification by projecting 3D regions onto multiple 2D views.
This approach requires both 2D images and 3D point clouds during inference. Additionally, it necessitates multiple inferences of large 2D models on projected masks, resulting in high computational costs. 
We address these limitations by developing the first single-stage open-vocabulary 3D instance segmentation model that operates directly in 3D without ground truth labels, using our \dataname dataset and Segment3D~\cite{huang2024segment3d} proposals.

\noindentbold{3D vision-language datasets}
Several datasets align 3D scenes with textual annotations to facilitate language-driven 3D understanding. 
ScanRefer~\cite{chen2020scanrefer}, ReferIt3D~\cite{achlioptas2020referit_3d} and EmbodiedScan~\cite{wangEmbodiedScanHolisticMultiModal2023} provide fine-grained object-level localization through detailed referential phrases, while ScanQA~\cite{azuma2022scanqa} targets spatially grounded question-answering. 
In contrast, SceneVerse~\cite{jiaSceneVerseScaling3D2024} and MMScan~\cite{lyu2024mmscan} employ large-language models or vision-language models to partially automate annotation.
Despite leveraging advanced models, these datasets depend significantly on costly human annotations derived from closed-vocabulary sources, limiting their support for open-vocabulary and scalability for large-scale 3D segmentation tasks.
}
\notoc{\section{\dataname Data Engine}
\label{sec:caption_generation}

Generating 3D mask-text pair datasets can be costly and require meticulous attention. Recent work~\cite{ding2022pla,yang2024regionplc,jiang2024open} have leveraged 2D visual foundation models (VFMs) to automate data annotation to an extent -- they use multi-view images to generate captions or features on different types of region proposals (e.g. bounding boxes, segmentations, or sliding windows). 
However, existing approaches suffer from imprecise boundary delineation due to their reliance on coarse object detectors~\cite{ding2022pla,yang2024regionplc}, 
or provide only simple attribute labels~\cite{jiang2024open}.
To overcome these limitations, we propose a data generation pipeline that combines recent advances in open-vocabulary segmentation and robust region-aware vision-language models (VLMs), 
enabling both precise region boundaries and rich descriptions that capture object attributes, spatial relationships, and scene context.

\begin{figure*}[t!]
    \centering
    \includegraphics[width=\linewidth]{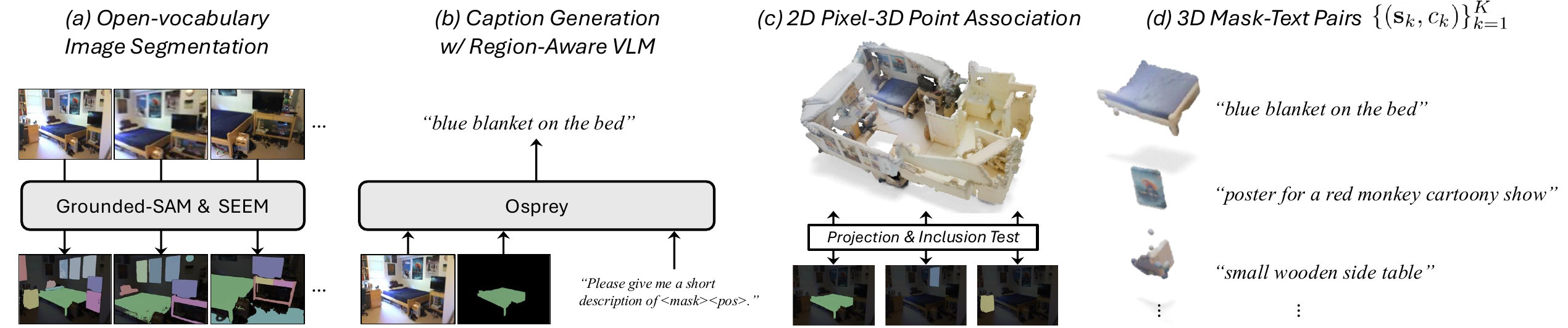}
    \caption{
        \textbf{\dataname data engine.} Our data generation process consists of three key steps: (a) We predict object segments for each RGB frame using state-of-the-art image segmentation models~\cite{sam,ravi2024sam,zou2024segment}. (b) We pass the images and predicted masks to a region-aware vision-language model~\cite{yuan2024osprey} to generate descriptive captions for each region. (c) We project the 2D segmentation masks onto 3D points using camera parameters to create (d) 3D mask-text pairs. This pipeline enables us to generate a large-scale dataset of 3D mask-text pairs.
    }
    \label{fig:data_pipeline}
\end{figure*}

\subsection{Proposed Pipeline}
Our pipeline overcomes limitations through two key improvements: accurate segmentation and the region captioning pipeline.
Fig.~\ref{fig:data_pipeline} illustrates the overview of our pipeline.

\noindent\textbf{Enhanced segmentation.}
We employ Grounded-SAM~\cite{ren2024grounded} and SEEM~\cite{zou2024segment} for more precise open-vocabulary image segmentation. We incorporate both models because Grounded-SAM excels at segmenting foreground objects with precise boundaries, while SEEM complements this with open-vocabulary panoptic segmentation that better handles background stuff like wall and floor. Thus, our method outperforms previous methods~\cite{yang2024regionplc,ding2022pla} that rely on foreground object detectors.

Given an RGB image $\mathbf{I}\in\mathbb{R}^{H \times W \times 3}$, Grounded-SAM first predicts open-set bounding boxes using Grounding-DINO~\cite{liu2023grounding}, then uses these boxes as input prompts for SAM~\cite{sam,ravi2024sam} to generate segmentation masks.
To fully automate the segmentation process, we employ RAM++~\cite{ram_pp} to detect object categories within the input image.
These detected categories serve as a text prompt for Grounding-DINO.
Through this, Grounded-SAM generates a set of segmentation masks $\{\mathbf{M}_k\}_{k=1}^K$
where $K$ is the number of detected objects in the image.
Each mask $\mathbf{M}_k \in \{0,1\}^{H \times W}$ represents a binary segmentation of an object.
SEEM operates in a similar way, except that it directly performs panoptic segmentation without being required to use RAM++ to generate tags or Grounding-DINO for object detection.
For simplicity, we denote the combined set of masks from both Grounded-SAM and SEEM as $\{\mathbf{M}_k\}_{k=1}^K$, where $K$ is the total number of masks from both models.

\noindent\textbf{Enhanced region captioning.}
For each segmentation mask, we generate a detailed caption that describes the visual characteristics and spatial context of the object.
Unlike previous methods that process an image as a whole using generic image captioning models~\cite{peng2023kosmos,llava,vit-gpt2,wang2022ofa}, we leverage region-aware vision-language models (VLMs)~\cite{you2023ferret,yuan2024osprey,rasheed2024glamm} that are specifically designed to understand and describe a region or a mask specified by a user as an additional. These models can generate detailed descriptions by interpreting various visual prompts, such as points, boxes, masks, and scribbles, enabling more focused and contextual captions for each segmented region.

Given an image $\mathbf{I}$, a segmentation mask $\mathbf{M}_k$, and a user prompt $\pi$ that asks for a detailed description of the masked region, the region-aware VLM $\mathcal{R}(\cdot)$ generates a textual response $c$ for each mask by
$c_k = \mathcal{R}(\mathbf{I}; \mathbf{M}_k, \pi)$,
where $c_k$ is a natural language description that captures the visual attributes and spatial context of the k-th masked region.
After evaluating several available region-aware VLMs, we chose Osprey~\cite{yuan2024osprey} for our implementation.

\noindent\textbf{2D pixel-3D point association.}
After obtaining segmentation masks and captions from multiple views, we associate them with 3D points to create mask-text pairs in 3D space.
For each 3D point $\mathbf{p}$ in the point cloud $\mathbf{P} \in \mathbb{R}^{N\times3}$, we project it onto each view using the camera parameters to obtain its 2D pixel coordinates $(u,v)$ and depth value $d$.
We check if the projected pixel falls within any segmentation mask $\mathbf{M}_k$ and if its depth matches the ground-truth depth at that location within a small threshold $\epsilon$ (inclusion test).
Specifically, for point cloud $\mathbf{P}$, 
we compute 3D binary region masks $\mathbf{s}_k \in \{0,1\}^N$ as:
\begin{equation}\label{eq:associate}
    \!\! \mathbf{s}_k = \left\{ 
    \begin{array}{cl}
        \! 1 & \!\!\! \textrm{if}~~(\mathbf{M}_{k})_{u,v} = 1 \textrm{ and } |d - \mathbf{D}_{u,v}|_1 < \epsilon \\
        \! 0 & \!\!\! \textrm{otherwise}
    \end{array}
    \right. \!\!\!\! ,
\end{equation}
where $\mathbf{D}$ is the ground-truth depth image. Finally, we obtain 3D mask-text pairs $\{(\mathbf{s}_k, c_k)\}_{k=1}^K$ that associate each segmented region with its corresponding caption.

Our pipeline produces high-quality 3D mask-text pairs that combine precise object boundaries with rich semantic descriptions. As demonstrated in Fig.~\ref{fig:data_comparison}, compared to previous methods, our approach achieves both more accurate segmentation boundaries and more detailed contextual descriptions that capture object attributes and spatial relationships.

\subsection{Data Statistics}
Given the limited availability of large-scale 3D scene datasets, it is crucial to leverage multiple existing datasets and apply a unified annotation process to create a comprehensive training corpus.
We curate a collection of widely-used 3D indoor scene datasets, including ScanNet~\cite{dai2017scannet}, ARKitScenes~\cite{baruch2021arkitscenes}, Matterport3D~\cite{chang2017matterport3d}, ScanNet++~\cite{yeshwanth2023scannet++}, and Structured3D~\cite{zheng2020structured3d}, and apply our proposed pipeline to each.
These datasets provide diverse indoor scenes covering both real and synthetic environments, with high-quality RGB-D scans, accurate camera poses, and dense 3D reconstructions, making them ideal for our automatic annotation process.

Through this data pipeline, we create \dataname, the largest 3D mask-text paired dataset to date,
encompassing over 30K indoor scenes and approximately 1M RGB-D frames, yielding 5.6M region captions comprising 30M total text tokens.
The complete data statistics can be found in 
Fig.~\ref{fig:data_stats}.
Our dataset offers significant advantages over the existing datasets in terms of:
\begin{itemize}[leftmargin=*,itemsep=1pt]
    \item \textbf{Scale:} We generate over 5.6M mask-text pairs with 30M text tokens across 30K scenes, significantly larger than previous datasets in scene coverage and annotation density.
    \item \textbf{Precision:} Our use of Grounded-SAM~\cite{ren2024grounded} and SEEM~\cite{zou2024segment} ensures precise region boundaries, significantly improving over bounding box-based approaches.
    \item \textbf{Richness:} The region-aware VLM generates detailed contextual descriptions that capture both visual attributes and spatial relationships, providing richer semantic information than simple object labels.
\end{itemize}

}

\notoc{\begin{figure*}[t!]
    \centering
    \includegraphics[width=0.9\linewidth]{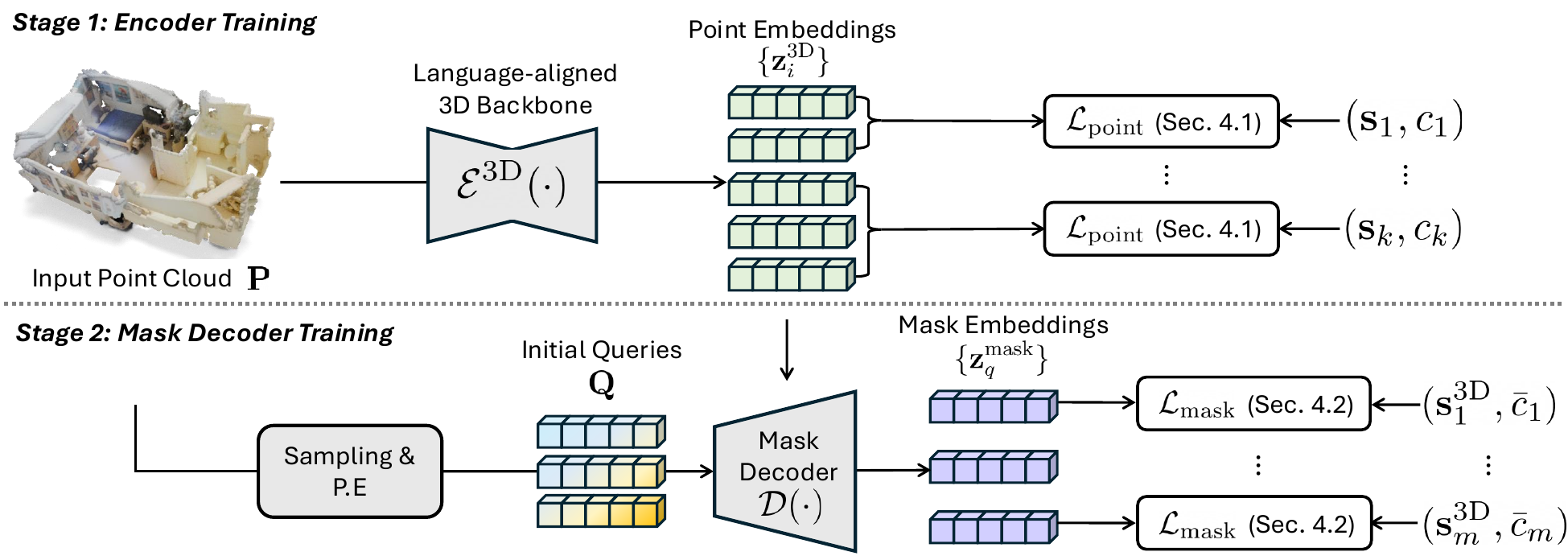}
    \caption{
        \textbf{\nickname model.}
        \nickname model is a SparseUNet~\cite{mink} trained with our \dataname dataset to extract language-aligned features from 3D point clouds.
        A mask decoder with positional encodings (P.E) is trained on top to enable instance segmentation.
    }
    \label{fig:architecture}
\end{figure*}

\section{\nickname Model Training}
We train the 3D open-vocabulary segmentation model based on \dataname using two-stage training: per-point language alignment (Sec.~\ref{subsec:backbone}), and mask decoder training that predicts instances from these aligned features (Sec.~\ref{subsec:mask_decoder}).

\subsection{\nickname: Language-Aligned 3D Encoder}
\label{subsec:backbone}

\noindentbold{Architecture}
We use U-shaped sparse convnets~\cite{graham20183d,mink} for our backbone due to their efficiency and scalability.
Given a point cloud $\mathbf{P} = \{\mathbf{p}_i\}_{i=1}^{N},\mathbf{p}_i \in \mathbb{R}^3$, the network $\mathcal{E}^{\text{3D}}(\cdot)$ outputs per-point features $\mathcal{E}^{\text{3D}}(\mathbf{P}) = \{\mathbf{z}^{\text{3D}}_i\}_{i=1}^N, \mathbf{z}^{\text{3D}}_i \in \mathbb{R}^D$.

\noindentbold{Training objective}
To align the geometric embeddings with language semantics, we employ a contrastive learning framework~\cite{yang2024regionplc,ding2022pla}.
Given 3D mask-text pairs $\{(\mathbf{s}_k, c_k)\}_{k=1}^K$, a pre-trained text encoder $\mathcal{E}^{\text{text}}(\cdot)$ computes text embeddings $\mathbf{z}^{\text{text}}_k \in \mathbb{R}^D$ for each caption.
The similarity scores between point features and text embeddings are averaged using the region masks to weigh all regions equally.
The final training objective is:
\begin{equation}
    \label{eq:caption_loss}
    \begin{aligned}
        \mathcal{L}_{\mathrm{point}} &= -\frac{1}{K}\sum_{k=1}^{K}\sum_{i=1}^{N}(\mathbf{s}_k)_i\log\frac{\exp(\mathbf{z}^{\text{3D}}_i\cdot\mathbf{z}^{\text{text}}_k/\tau)}{\sum_{j=1}^{K}\exp(\mathbf{z}^{\text{3D}}_i\cdot\mathbf{z}^{\text{text}}_j)}
    \end{aligned}
\end{equation}
where $\tau$ is a learnable logit temperature.

\subsection{\nickname with Mask Decoder}
\label{subsec:mask_decoder}
\begin{algorithm}[t]
\caption{Caption Merging}
\label{alg:caption_merging}
\begin{algorithmic}[1]
\Require Our mask-text pairs data $\{(\mathbf{s}_k, c_k)\}_{k=1}^K$, Segment3D~\cite{huang2024segment3d} masks $\{\mathbf{s}_{l}^{\mathrm{3D}}\}_{l=1}^L$
\Ensure Matched mask-text pairs $\{(\mathbf{s}^{\mathrm{3D}}_m, \{c_{j\in \mathcal{M}_m}\})\}_{m=1}^M$
\For{$k \gets 1$ to $K$}
    \State $l^* = \arg\max_l \mathrm{IoU}(\mathbf{s}_k, \mathbf{s}_{l}^{\mathrm{3D}})$ \Comment{Find match}
    \If{$\mathcal{M}_{l^*}$ is not initialized}
        \State $\mathcal{M}_{l^*} \gets \emptyset$ \Comment{Initialize empty set}
    \EndIf
    \If{$\mathrm{IoU}(\mathbf{s}_k, \mathbf{s}_{l^*}^{\mathrm{3D}}) > \tau$}
        \State $\mathcal{M}_{l^*} \gets \mathcal{M}_{l^*} \cup \{k\}$ \Comment{Add caption}
    \EndIf
\EndFor
\State \Return $\{(\mathbf{s}^{\mathrm{3D}}_m, \{c_{j\in \mathcal{M}_m}\})\}_{m=1}^M$
\end{algorithmic}
\end{algorithm}

On top of our language-aligned backbone, we add a lightweight mask decoder to enable open-vocabulary 3D instance segmentation, avoiding the need for separate instance segmentation networks used in prior work~\cite{takmaz2023openmask3d,nguyen2024open3dis,huang2024openins3d,yin2024sai3d}.

\noindentbold{Architecture}
We use Mask3D~\cite{schult2023mask3d} as our mask decoder, a transformer-based architecture adapted from 2D segmentation~\cite{cheng2021per,cheng2022masked}.
Specifically, our decoder takes non-parametric queries (\ie, positional encodings of points sampled from the input point cloud) and language-aligned point features from our backbone (Sec.~\ref{subsec:backbone}) as input.
The decoder outputs mask embeddings aligned with language features, enabling open-vocabulary segmentation of 3D scenes.

\noindentbold{Training data}
To enable open-vocabulary 3D instance segmentation, we need training data that is not constrained to predefined categories.
While prior work~\cite{takmaz2023openmask3d,nguyen2024open3dis,huang2024openins3d} used closed-vocabulary labels, we leverage Segment3D~\cite{huang2024segment3d}'s class-agnostic masks predicted by SAM~\cite{sam} and combine them with our multi-view mask-caption data to create a rich open-vocabulary training set.
As detailed in Algorithm~\ref{alg:caption_merging}, we merge our mask-caption data $\{(\mathbf{s}_k, c_k)\}_{k=1}^K$ with Segment3D~\cite{huang2024segment3d} masks $\{\mathbf{s}_{l}^{\mathrm{3D}}\}_{l=1}^L$ based on IoU matching.
This yields a set of Segment3D masks $\{(\mathbf{s}^{\mathrm{3D}}_m, \{c_{j\in \mathcal{M}_m}\})\}_{m=1}^M$, each associated with multiple captions from our multi-view data.
During training, we randomly sample a fixed number of captions for each mask from its associated caption set.

\noindentbold{Training objective}
Given a point cloud $\mathbf{P}$ and Segment3D masks with associated captions $\{(\mathbf{s}^{\mathrm{3D}}_m, \{c_{j \in \mathcal{M}_m}\})\}_{m=1}^M$, we compute $Q$ numbers of mask embeddings $\mathbf{Z}^{\mathrm{mask}} \in \mathbb{R}^{Q \times D}$ and normalized text embeddings $\bar{\mathbf{z}}^{\text{text}}_m \in \mathbb{R}^D$:
\begin{equation}
    \label{eq:mask_decoder}
    \mathbf{Z}^{\mathrm{mask}} = \{\mathbf{z}^{\text{mask}}_q\} = \mathcal{D}(\{\mathbf{z}^{\text{3D}}_i\}; \mathbf{Q}), \quad \mathbf{\bar{z}}^{\text{text}}_m = \mathcal{E}^{\text{text}}(\bar{c}_m),
\end{equation}
where $\mathcal{D}(\cdot)$ is our mask decoder that takes point features and sampled queries $\mathbf{Q}$ as input, and $\bar{c}_m$ concatenates all captions associated with mask $\mathbf{s}^{\mathrm{3D}}_m$.
Following Segment3D~\cite{huang2024segment3d}, we first train the mask decoder to predict binary instance masks using three standard losses: objectness prediction loss $\mathcal{L}_{\mathrm{obj}}$, Dice loss $\mathcal{L}_{\mathrm{dice}}$~\cite{dice_loss}, and binary cross entropy loss $\mathcal{L}_{\mathrm{bce}}$. Then, we use Hungarian matching to find the set of mask predictions that minimizes error given ground-truth masks.
The losses are computed as:
\begin{equation}
    \label{eq:mask_prediction}
    \mathbf{o} = \mathrm{Linear}(\mathbf{Z}^{\mathrm{mask}}), \quad \mathbf{S} = \sigma(\mathbf{Z}
    ^{\mathrm{mask}} \cdot \mathbf{Z}^{\text{3D}\top}),
\end{equation}
where $\mathbf{o} \in \mathbb{R}^{Q \times 2}$ and $\mathbf{S} \in \mathbb{R}^{Q \times N}$ are objectness scores and predicted binary masks.
To enable open-vocabulary segmentation, we introduce a mask caption loss $\mathcal{L}_{\mathrm{cap}}$ that explicitly aligns mask embeddings with caption embeddings:
\begin{equation}
    \label{eq:mask_caption_loss}
    \mathcal{L}_{\mathrm{cap}} = -\frac{1}{M}\sum_{m=1}^{M}\log\frac{\exp(\mathbf{z}^{\text{mask}}_m\cdot\bar{\mathbf{z}}^{\text{text}}_k/\tau)}{\sum_{j=1}^{M}\exp(\mathbf{z}^{\text{mask}}_m\cdot\bar{\mathbf{z}}^{\text{text}}_j)}
\end{equation}
The total loss is $\mathcal{L}_{\mathrm{mask}} = \lambda_{\mathrm{obj}}\mathcal{L}_{\mathrm{obj}} + \lambda_{\mathrm{dice}}\mathcal{L}_{\mathrm{dice}} + \lambda_{\mathrm{bce}}\mathcal{L}_{\mathrm{bce}} + \lambda_{\mathrm{cap}}\mathcal{L}_{\mathrm{cap}}$, where in practice we set $\lambda_{\mathrm{obj}}=2$, $\lambda_{\mathrm{dice}}=5$, $\lambda_{\mathrm{bce}}=2$, and $\lambda_{\mathrm{cap}}=1$.
With this loss, our language-aligned mask decoder enables direct open-vocabulary 3D segmentation, avoiding the expensive multi-view CLIP inference required by prior methods~\cite{takmaz2023openmask3d,nguyen2024open3dis,huang2024openins3d}.
}
\notoc{\section{Experiments}
\label{sec:experiments}
In this section, we investigate how dataset size impacts model performance (Sec.~\ref{subsec:scaling}).
Next, we benchmark \nickname against existing methods for open-vocabulary 3D scene segmentation (Sec.~\ref{subsec:benchmark}).
Finally, we analyze our system through attention visualization, zero-shot experiments, and ablation studies on the \dataname data engine (Sec.~\ref{subsec:analysis}).

\subsection{Setup}

\noindentbold{Implementation details}
We adopt Sparse ConvNets~\cite{graham20183d,mink} as our 3D encoder, leveraging their efficiency in processing sparse 3D data.%
Our baseline architecture uses SparseUNet34C~\cite{mink} with 43.7M trainable parameters. %
For text encoder, we employ Recap-CLIP~\cite{li2024if}, which is pre-trained on longer re-captioned datasets, enabling better processing of the long captions in our dataset.
For training, we use SGD optimization~\cite{bottou2010large} with an initial learning rate of $0.05$ and a weight decay of $1\times10^{-4}$, coupled with the OneCycleLR scheduler~\cite{smith2019super}, with batch size 4. 
To enhance multi-data joint training, we adopt recent technique of Point Prompt Training (PPT)~\cite{wu2023towards}.
All models are trained for 128 epochs on eight A100 GPUs.
For instance segmentation, we fine-tune the pre-trained \nickname model with an additional mask decoder on \dataname with caption merging (Alg.~\ref{alg:caption_merging}).
Please refer to the appendix for more details.

\noindentbold{Evaluation metrics}
We evaluate performance using mean Intersection over Union (mIoU) and mean Accuracy (mAcc), which are standard metrics for open-vocabulary 3D semantic segmentation. 
Following prior work~\cite{jiang2024open,yang2024regionplc,ding2022pla}, we report f-mIoU and f-mAcc metrics that exclude background classes.

\subsection{Impact of Dataset Size}
\label{subsec:scaling}

\begin{figure}[t]
    \centering
    \includegraphics[width=0.95\linewidth]{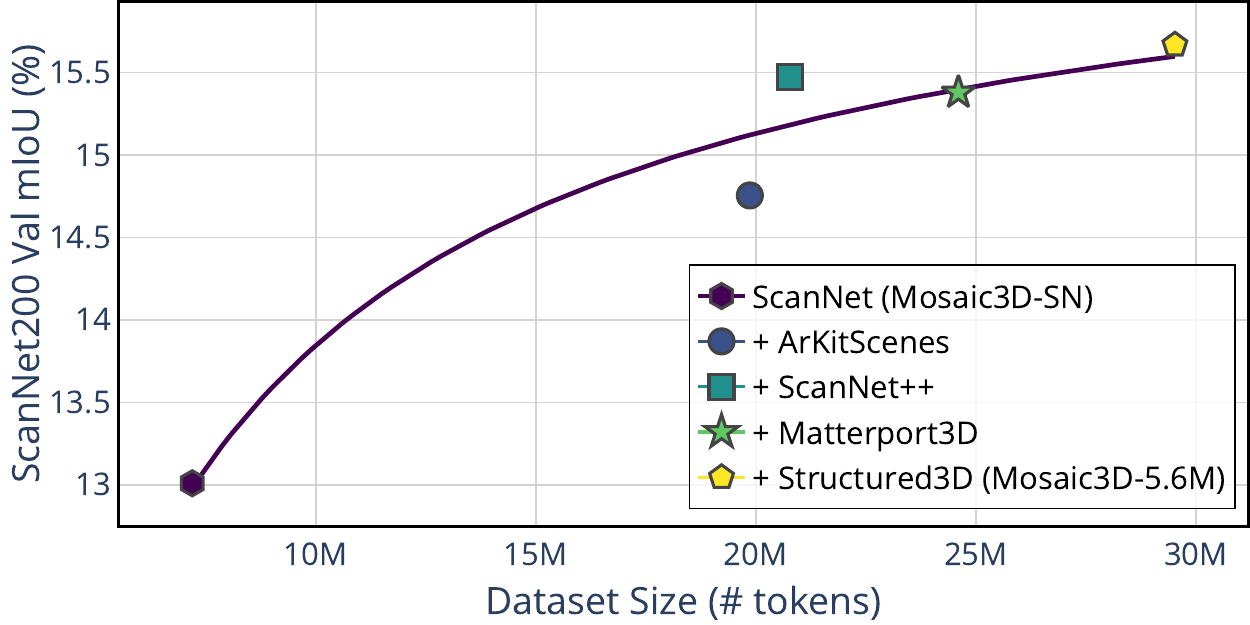}
    \vspace{-2mm}
    \caption{
        \textbf{Model performance scales with training data.}
        We observe consistent improvements in open-vocabulary semantic segmentation on ScanNet200~\cite{scannet200} as we increase the amount of training data.
        This shows the value of our large-scale data generation pipeline in improving open-vocabulary 3D scene understanding.
    }
    \label{fig:scaling}
\end{figure}

\begin{table*}[!t]
    \centering
        \resizebox{\linewidth}{!}{
        \begin{tabular}{ll|rr|rr|rr|rr}
            \toprule
            \multirow{2}{*}{Method} & \multirow{2}{*}{Source Datasets} & \multicolumn{2}{c|}{ScanNet20 (20)} & \multicolumn{2}{c|}{ScanNet++ (100)} & \multicolumn{2}{c|}{Matterport3D (160)} & \multicolumn{2}{c}{ScanNet200 (200)} \\
            &  & f-mIoU & f-mAcc & f-mIoU & f-mAcc & f-mIoU & f-mAcc & f-mIoU & f-mAcc \\
            \midrule
            OpenScene-3D$^{\dagger}$~\cite{Peng2023OpenScene} & ScanNet~\cite{dai2017scannet} & 57.5 & 72.4 & 8.8 & 14.7 & 5.7 & 10.7 & 6.4 & 12.2 \\
            PLA~\cite{ding2022pla} & ScanNet~\cite{dai2017scannet} & 19.1 & 41.5 & - & - & - & - & 1.8 & 3.1 \\
            RegionPLC~\cite{yang2024regionplc} & ScanNet~\cite{dai2017scannet} & 59.6 & 77.5 & - & - & - & - & 9.1 & 17.3 \\
            RegionPLC$^{\flat}$~\cite{yang2024regionplc} & ScanNet~\cite{dai2017scannet} & 55.6 & 76.3 & 11.3 & 20.1 & 6.2 & 13.3 & 9.2 & 16.4 \\
            OV3D~\cite{jiang2024open} & ScanNet~\cite{dai2017scannet} & 64.0 & 76.3 & - & - & - & - & 8.7 & - \\
            \rowcolor{gray!15} \nickname & ScanNet~\cite{dai2017scannet} & \textbf{65.0} & \textbf{82.5} & \textbf{16.2} & \textbf{27.1} & \textbf{8.6} & \textbf{17.8} & \textbf{13.0} & \textbf{24.5} \\
            \midrule
            \multirow{2}{*}{\shortstack{RegionPLC~\cite{yang2024regionplc} \\ + SceneVerse~\cite{jiaSceneVerseScaling3D2024}}} & \multirow{2}{*}{\shortstack{MS~\cite{mao2022multiscan} + 3RS~\cite{wald2019rio} + SN~\cite{dai2017scannet} + AR~\cite{baruch2021arkitscenes} \\ + HM3D~\cite{ramakrishnan2021habitat} + S3D~\cite{zheng2020structured3d} + PT~\cite{deitke2022️}}} & \multirow{2}{*}{61.0} & \multirow{2}{*}{79.7} & \multirow{2}{*}{-} & \multirow{2}{*}{-} & \multirow{2}{*}{-} & \multirow{2}{*}{-} & \multirow{2}{*}{-} & \multirow{2}{*}{-} \\
            & & & & & & & & \\
            \rowcolor{gray!15} \nickname & SN~\cite{dai2017scannet} + AR~\cite{baruch2021arkitscenes} + SN2~\cite{yeshwanth2023scannet++} + M~\cite{chang2017matterport3d} + S3D~\cite{zheng2020structured3d} & \textbf{68.1} & \textbf{84.4} & \textbf{18.0} & \textbf{29.0} & \textbf{13.1} & \textbf{27.7} & \textbf{15.7} & \textbf{28.3} \\
            \bottomrule
        \end{tabular}
        }
        \vspace{-2mm}
        \caption{
            \textbf{Annotation-free 3D semantic segmentation on ScanNet~\cite{dai2017scannet,scannet200}, Matterport3D~\cite{chang2017matterport3d}, and ScanNet++~\cite{yeshwanth2023scannet++}.}
            We report f-mIoU and f-mAcc excluding background classes (wall, floor, ceiling), following~\cite{yang2024regionplc,Peng2023OpenScene,ding2022pla,jiaSceneVerseScaling3D2024}.
            $^{\dagger}$ denotes official checkpoints and $^{\flat}$ denotes our reproductions.
            The numbers in parentheses indicate the total number of classes in each dataset.
            Dataset abbreviations SN, AR, SN2, M, S3D, MS, 3RS, HM3D, and PT denote ScanNet~\cite{dai2017scannet}, ARKitScenes~\cite{baruch2021arkitscenes}, ScanNet++~\cite{yeshwanth2023scannet++}, Matterport3D~\cite{chang2017matterport3d}, Structured3D~\cite{zheng2020structured3d}, MultiScan~\cite{mao2022multiscan}, 3RScan~\cite{wald2019rio}, Habitat-Matterport 3D~\cite{ramakrishnan2021habitat}, and ProcTHOR~\cite{deitke2022️}, respectively.
        }
    \label{tab:scannet200_semantic}
\end{table*}

\begin{table*}[!t]
    \centering
    \resizebox{\linewidth}{!}{
        \begin{tabular}{llllrrrrrrr}
            \toprule
            & Method & Inputs & 3D Region Proposal Network & mAP & mAP$_{50}$ & mAP$_{25}$ & mAP$_{\mathrm{head}}$ & mAP$_{\mathrm{com.}}$ & mAP$_{\mathrm{tail}}$ & Latency \\
            \midrule
            \multirow{3}{*}{(a)} & \multirow{2}{*}{Open3DIS~\cite{nguyen2024open3dis}} & \multirow{2}{*}{3D + 2D} & Superpoints~\cite{felzenszwalb2004efficient} + ISBNet~\cite{ngo2023isbnet} & \multirow{2}{*}{23.7} & \multirow{2}{*}{29.4} & \multirow{2}{*}{32.8} & \multirow{2}{*}{27.8} & \multirow{2}{*}{21.2} & \multirow{2}{*}{21.8} & \multirow{2}{*}{33.5} \\
            & & & + Grounded-SAM~\cite{ren2024grounded} & & & & & & & \\
            & SAI3D~\cite{yin2024sai3d} & 3D + 2D & Superpoints~\cite{felzenszwalb2004efficient} + SAM~\cite{sam} & 12.7 & 18.8 & 24.1 & 12.1 & 10.4 & 16.2 & 75.2 \\
            \midrule
            \multirow{3}{*}{(b)} & OpenScene-2D~\cite{Peng2023OpenScene} & 3D + 2D &  Mask3D~\cite{schult2023mask3d} & 11.7 & 15.2 & 17.8 & 13.4 & 11.6 & 9.9 & - \\
            & OpenScene-2D/3D~\cite{Peng2023OpenScene} & 3D + 2D & Mask3D~\cite{schult2023mask3d} & 5.3 & 6.7 & 8.1 & 11.0 & 3.2 & 1.1 & - \\
            & OpenMask3D~\cite{takmaz2023openmask3d} & 3D + 2D &  Mask3D~\cite{schult2023mask3d} & 15.4 & 19.9 & 23.1 & 17.1 & 14.1 & 14.9 & 47.3 \\
            \midrule
            \multirow{5}{*}{(c)} & OpenScene-3D~\cite{Peng2023OpenScene} & 3D & Mask3D~\cite{schult2023mask3d} & 4.8 & 6.2 & 7.2 & 10.6 & 2.6 & 0.7 & 1.1 \\
            & RegionPLC~\cite{yang2024regionplc} & 3D & Mask3D~\cite{schult2023mask3d} & 6.3 & 8.6 & 9.7 & 15.6 & 1.0 & 1.7 & \textbf{1.0} \\ %
            & OpenIns3D~\cite{huang2024openins3d} & 3D & Mask3D~\cite{schult2023mask3d} & 8.8 & 10.3 & 14.4 & 16.0 & 6.5 & 4.2 & 285.2 \\
            & OpenIns3D$^{\dagger}$~\cite{huang2024openins3d} & 3D & Mask3D~\cite{schult2023mask3d} & 3.3 & 5.0 & 5.6 & 7.0 & 1.4 & 1.2 & 50.0 \\
            & \cellcolor{gray!15}\nickname & \cellcolor{gray!15}3D & \cellcolor{gray!15}Mask3D~\cite{schult2023mask3d} & \cellcolor{gray!15}\textbf{11.8} & \cellcolor{gray!15}\textbf{16.0} & \cellcolor{gray!15}\textbf{17.8} & \cellcolor{gray!15}\textbf{21.8} & \cellcolor{gray!15}\textbf{7.2} & \cellcolor{gray!15}\textbf{5.4} & \cellcolor{gray!15}\textbf{1.0} \\ 
            \midrule
            \multirow{5}{*}{(d)} & OpenScene-3D~\cite{Peng2023OpenScene} & 3D & Segment3D~\cite{huang2024segment3d} & 0.6 & 1.0 & 1.6 & 1.4 & 0.4 & 0.0 & 2.0 \\
            & RegionPLC~\cite{yang2024regionplc} & 3D & Segment3D~\cite{huang2024segment3d} & 1.5 & 2.1 & 2.6 & 2.3 & 0.2 & 1.9 & 1.9 \\ %
            & OpenIns3D$^{\dagger}$~\cite{huang2024openins3d} & 3D & Segment3D~\cite{huang2024segment3d} & 1.7 & 2.7 & 3.7 & 3.2 & 0.8 & 1.0 & 64.8 \\
            & \cellcolor{gray!15}\nickname & \cellcolor{gray!15}3D & \cellcolor{gray!15}Segment3D~\cite{huang2024segment3d} & \cellcolor{gray!15}2.7 & \cellcolor{gray!15}4.2 & \cellcolor{gray!15}5.7 & \cellcolor{gray!15}3.8 & \cellcolor{gray!15}2.0 & \cellcolor{gray!15}2.4 & \cellcolor{gray!15}1.9 \\
            & \cellcolor{gray!15}\nickname w/ Decoder & \cellcolor{gray!15}3D & \cellcolor{gray!15}\xmark & \cellcolor{gray!15}\textbf{3.9} & \cellcolor{gray!15}\textbf{7.0} & \cellcolor{gray!15}\textbf{12.3} & \cellcolor{gray!15}\textbf{6.6} & \cellcolor{gray!15}\textbf{2.1} & \cellcolor{gray!15}\textbf{2.8} & \cellcolor{gray!15}\textbf{1.2}\\
            \bottomrule
        \end{tabular}
    }
    \vspace{-2mm}
    \caption{
        \textbf{Annotation-free 3D instance segmentation on ScanNet200~\cite{scannet200}.}
        For a fair comparison, we categorize methods by input types and region proposal network:
        \textbf{(a)} Methods using both 3D point cloud and 2D RGB-D images, with 3D+2D region proposals and 2D CLIP inference.
        \textbf{(b)} Methods using both 3D+2D inputs, with region proposals from Mask3D~\cite{schult2023mask3d} (closed-vocab) and 2D CLIP inference.
        \textbf{(c)} Methods using only 3D input with Mask3D~\cite{schult2023mask3d}.
        \textbf{(d)} Methods using only 3D input with open-vocabulary 3D region proposals.
        $^{\dagger}$ denotes results without test-time voting, following the official implementation.
        Latency reports runtime (seconds) per scene on ScanNet validation.
    }
    \label{tab:scannet200_instance}
\end{table*}

To understand how the size of the training data affects model performance, we conduct experiments with varying amounts of training data.
Specifically, we gradually increase the training data by adding one dataset at a time in the following order: ScanNet~\cite{dai2017scannet}, ARKitScenes~\cite{baruch2021arkitscenes}, ScanNet++~\cite{yeshwanth2023scannet++}, Matterport3D~\cite{chang2017matterport3d}, and Structured3D~\cite{zheng2020structured3d}.
For these experiments, we fix the model architecture to SparseUNet34C~\cite{mink}.
All other hyperparameters remain fixed across the experiments.
As shown in Fig.~\ref{fig:scaling}, increasing the size of the dataset generally improves the accuracy of open-vocabulary semantic segmentation on the ScanNet200 benchmark~\cite{scannet200}.
The most significant performance gains are observed when incorporating ARKitScenes and ScanNet++, which we attribute to their high-quality, dense RGB-D frames captured from real 3D environments. 
From here on, we refer to our \nickname model as the model jointly trained on all datasets.

\subsection{Benchmark Results}
\label{subsec:benchmark}

\noindentbold{Open-vocabulary 3D semantic segmentation}
We evaluate on ScanNet20 validation set, ScanNet200 validation set, Matterport3D test set, and ScanNet++ validation set, following prior work~\cite{Peng2023OpenScene,ding2022pla,yang2024regionplc,jiang2024open}.
These datasets contain 20, 200, 160, and 100 semantic classes respectively, providing diverse benchmarks for open-vocabulary 3D semantic segmentation.
Using only ScanNet training data, \nickname outperforms prior work in terms of f-mIoU (\%) on all benchmarks: surpassing OV3D~\cite{jiang2024open} by 1.0p on ScanNet20 and RegionPLC~\cite{yang2024regionplc} by 4.9p, 2.4p, and 3.8p on ScanNet++, Matterport3D, and ScanNet200 respectively.
Training on our full dataset \dataname further improves f-mIoU (\%) across all benchmarks, achieving 68.1 on ScanNet20, 18.0 on ScanNet++, 13.1 on Matterport3D, and 15.7 on ScanNet200.
Notably, our approach achieves 7.1p higher f-mIoU than SceneVerse~\cite{jiaSceneVerseScaling3D2024} despite using fewer scenes, highlighting the importance of caption quantity per scene.

\noindentbold{Open-vocabulary 3D instance segmentation}
As shown in Table~\ref{tab:scannet200_instance}, methods using both 2D and 3D inputs achieve strong results by directly applying CLIP models, but are impractical due to high latency (33-285 sec per scene) from processing multiple view images.
For fair comparison, we evaluate our \nickname trained only on ScanNet against prior methods~\cite{huang2024openins3d,takmaz2023openmask3d,yin2024sai3d,nguyen2024open3dis,Peng2023OpenScene,yang2024regionplc} that also train and evaluate on ScanNet.
With Mask3D~\cite{schult2023mask3d} as a closed-vocabulary region proposal network, \nickname outperforms the previous best method OpenIns3D~\cite{huang2024openins3d} by 3.0p mAP.
When using truly open-vocabulary Segment3D~\cite{huang2024segment3d} proposals, \nickname maintains strong performance at 2.7 mAP despite the more challenging setting.
Finally, our lightweight mask decoder trained with \dataname with caption merging (Alg.~\ref{alg:caption_merging}) achieves 3.9 mAP while being the first single-stage open-vocabulary 3D instance segmentation model that does not require ground truth labels.

\subsection{Analysis}
\label{subsec:analysis}
\begin{table*}[!t]
    \centering
        \resizebox{\linewidth}{!}{
        \begin{tabular}{llr|rrrr|rrrr}
            \toprule
            \multirow{2}{*}{Segmentation} & \multirow{2}{*}{Captioning} & \multirow{2}{*}{\# frames (K)} & \multicolumn{4}{c|}{ScanNet20~\cite{dai2017scannet}} & \multicolumn{4}{c}{ScanNet200~\cite{scannet200}} \\
            & & & mIoU & mAcc & f-mIoU & f-mAcc & mIoU & mAcc & f-mIoU & f-mAcc \\
            \midrule
            Detic~\cite{zhou2022detecting} & Kosmos-2~\cite{peng2023kosmos} & 125 & 32.7 & 64.1 & 52.3 & 73.2 & 6.2 & 14.1 & 7.4 & 14.2 \\
            LLaVA-1.5~\cite{liu2024improved} + SEEM~\cite{zou2024segment} & LLaVA-1.5~\cite{liu2024improved} & 25 & 30.1 & 61.9 & 45.9 & 68.1 & 4.6 & 13.0 & 5.7 & 13.0 \\
            RAM++~\cite{ram_pp} + SEEM~\cite{zou2024segment} & LLaVA-1.5~\cite{liu2024improved} & 25 & 41.3 & 67.1 & 57.0 & 74.6 & 6.8 & 13.2 & 7.4 & 13.2 \\
            RAM++~\cite{ram_pp} + Grounded-SAM~\cite{ren2024grounded,sam} & Ferret~\cite{you2023ferret} & 25 & 41.9 & 71.1 & 59.6 & 79.2 & 8.2 & 17.8 & 9.0 & 17.8 \\
            RAM++~\cite{ram_pp} + Grounded-SAM~\cite{ren2024grounded,sam} & Osprey~\cite{yuan2024osprey} & 25 & 46.2 & 72.0 & 63.7 & 80.7 & 8.4 & 18.2 & 9.2 & 18.2 \\
            RAM++~\cite{ram_pp} + Grounded-SAM2~\cite{ren2024grounded,ravi2024sam} & Osprey~\cite{yuan2024osprey} & 25 & 45.1 & 71.3 & 62.3 & 79.7 & 9.5 & 20.0 & 10.6 & \underline{20.3} \\
            RAM++~\cite{ram_pp} + Grounded-SAM2~\cite{ren2024grounded,ravi2024sam} & Osprey~\cite{yuan2024osprey} & 125 & 46.1 & \underline{73.0} & \underline{65.0} & \underline{81.6} & \underline{10.2} & \textbf{21.2} & \textbf{11.5} & \textbf{21.3} \\
            \rowcolor{gray!15} RAM++~\cite{ram_pp} + Grounded-SAM2~\cite{ren2024grounded,ravi2024sam} \& SEEM~\cite{zou2024segment} & Osprey~\cite{yuan2024osprey} & 125 & \textbf{50.0} & \textbf{73.7} & \textbf{65.2} & \textbf{82.0} & \textbf{10.5} & \underline{20.2} & \underline{11.0} & 20.1 \\
            \bottomrule
        \end{tabular}
        }
        \vspace{-2mm}
        \caption{
            \textbf{Data pipeline comparison.}
            We evaluate data generation pipelines for annotation-free 3D semantic segmentation on ScanNet20~\cite{dai2017scannet} and ScanNet200~\cite{scannet200}.
            All experiments use RegionPLC's~\cite{yang2024regionplc} architecture and training objective.
        }
    \label{tab:pipeline_comparison}
\end{table*}

\begin{figure*}[t!]
    \centering
    \includegraphics[width=\linewidth]{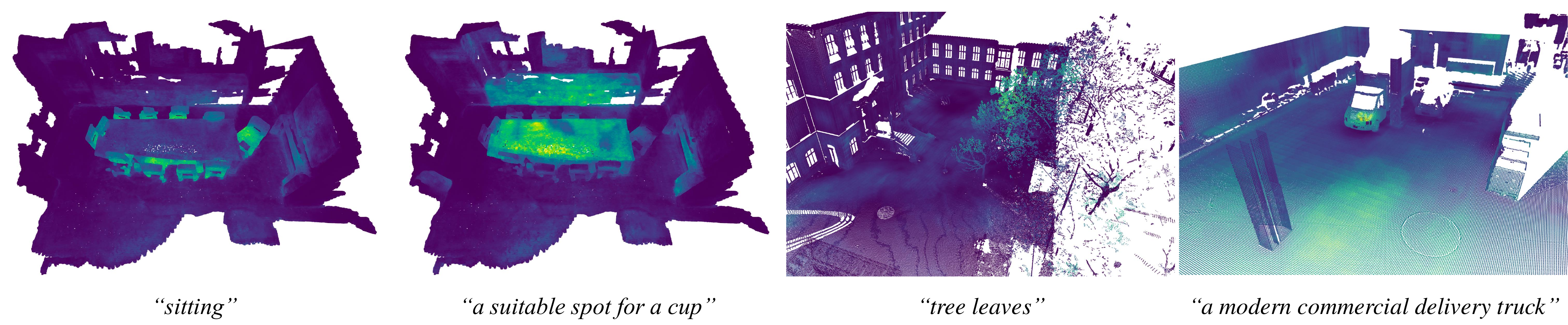}
    \vspace{-7mm}
    \caption{
        \textbf{Attention visualization of \nickname as a 3D foundational model.}
        From left to right, the first two examples show results on ScanNet~\cite{dai2017scannet}, while the next two examples show results on ETH3D~\cite{schops2017multi}.
        More examples are provided in the supplmentary material.
    }
    \label{fig:qual}
\end{figure*}

\noindentbold{Data engine components}
Finally, we conduct an ablation study on different component combinations of \nickname data engine to measure the contribution of each component. 
To understand the impact of different components in our data generation pipeline, we conduct ablation experiments by systematically varying key components while keeping the model architecture (SparseUNet16~\cite{yang2024regionplc}) and source dataset (ScanNet only) fixed.
As shown in Table~\ref{tab:pipeline_comparison}, we evaluate the following configurations:

\begin{itemize}[leftmargin=*,itemsep=1pt]
    \item \textbf{Mask Generation:} We compare our Grounded-SAM~\cite{ren2024grounded} + SEEM~\cite{zou2024segment} approach against using only Grounded-SAM~\cite{ren2024grounded} or SEEM~\cite{zou2024segment}. Results show that combining both methods leads to better region proposals.
    \item \textbf{Caption Generation:} We compare LLaVA~\cite{llava} for image-level captioning against Ferret~\cite{you2023ferret} and Osprey~\cite{yuan2024osprey} for region-level captioning. Region-level approaches perform better by providing detailed per-region descriptions, with Osprey achieving the best results.
\end{itemize}
Our final pipeline combines Grounded-SAM2~\cite{ren2024grounded,ravi2024sam} with SEEM~\cite{zou2024segment} for mask generation and Osprey~\cite{yuan2024osprey} for captioning, as this configuration yields optimal performance.

\noindentbold{Attention visualization}
In Fig.~\ref{fig:qual}, we visualize the similarity between dense point features obtained from \nickname and free-form text queries.
As shown in the figure, despite being trained only on indoor scenes, \nickname effectively localizes semantic regions described by text queries across both indoor~\cite{dai2017scannet} and outdoor~\cite{schops2017multi} datasets.

\noindentbold{Zero-shot 3D semantic segmentation}
While annotation-free methods~\cite{yang2024regionplc,jiang2024open} do not utilize GT annotations, their generated captions often contain class names from the evaluation set (\eg, "chair"), making them not truly "zero-shot."
To address this, we conduct a more rigorous zero-shot analysis by anonymizing all class names in the training captions, replacing them with general terms like "object."
To ensure fair comparison, we kept all experimental settings (\eg, model architecture, CLIP model, loss function) identical across methods, varying only the training data.
As shown in Table~\ref{tab:anonymization}, all models experience performance drops when trained on anonymized data, but the model trained on \dataname maintains the strongest performance (10.5 f-mIoU), outperforming both annotation-free methods and approaches that use ground truth annotations like LEO~\cite{huangEmbodiedGeneralistAgent2023}.
\begin{table}[!t]
    \centering
        \resizebox{0.72\linewidth}{!}{
        \begin{tabular}{lrrr}
            \toprule
            \multirow{2}{*}{Dataset} & \multicolumn{2}{c}{f-mIoU} & \multirow{2}{*}{$\Delta$} \\
            & w/o Anon. & w/ Anon. \\
            \midrule
            RegionPLC~\cite{yang2024regionplc} & 8.5 & 3.8 & $\downarrow$~~~4.7\\
            OV3D$^{\flat}$~\cite{jiang2024open} & 9.2 & 4.3 & $\downarrow$~~~4.9\\
            \rowcolor{gray!15}Mosaic3D-SN\tablefootnote{A subset of Mosaic3D-5.6M using only ScanNet as source dataset} & \textbf{13.0} & \textbf{8.7} & $\downarrow$~~~4.3\\
            \midrule
            LEO~\cite{huangEmbodiedGeneralistAgent2023} & 14.8 & 2.9 & $\downarrow$ 11.9\\
            SceneVerse~\cite{jiaSceneVerseScaling3D2024} & 13.6 & 4.7 & $\downarrow$~~~8.9\\
            EmbodiedScan~\cite{wangEmbodiedScanHolisticMultiModal2023} & 6.7 & 0.0 & $\downarrow$~~~6.7\\
            MMScan~\cite{lyu2024mmscan} & 11.7 & 8.5 & $\downarrow$~~~3.2\\
            \rowcolor{gray!15}Mosaic3D-5.6M & \textbf{15.7} & \textbf{10.5} & $\downarrow$~~~5.2\\
            \bottomrule
        \end{tabular}
        }
    \vspace{-2mm}
    \caption{
        \textbf{Zero-shot semantic segmentation on ScanNet200~\cite{scannet200}}.
        ``Anon.'' refers to class name anonymization, and $\flat$ indicates our re-implementation of OV3D as the original data is not available.
    }
    \label{tab:anonymization}
\end{table}

}
\notoc{\section{Conclusion}
\label{sec:conclusion}

In this work, we introduced a comprehensive approach for open-vocabulary 3D scene understanding that addresses fundamental data and modeling challenges in the field.
Our key contribution is a novel dataset generation pipeline that leverages state-of-the-art 2D visual foundation models to create high-quality 3D mask-text pairs, enabling the creation of \dataname, the largest open-vocabulary 3D scene dataset to date with 5.6M captions.
Building on this data, we developed a model that combines \nickname, a language-aligned 3D encoder, with a lightweight mask decoder, achieving state-of-the-art results on open-vocabulary 3D segmentation tasks.
Our ablation studies demonstrate the importance of dataset scale and annotation quality for open-vocabulary 3D understanding, providing a foundation for leveraging 2D vision models in 3D scene understanding.

\noindentbold{Acknowledgement}
This work was partly supported by the Institute of Information \& Communications Technology Planning \& Evaluation (IITP) grants (RS-2021-II212068: AI Innovation Hub, RS-2024-00457882: National AI Research Lab Project) funded by the Korea government (MSIT).
}

{
    \small
    \bibliographystyle{ieeenat_fullname}
    \bibliography{main}
}

\clearpage
\appendix
\maketitlesupplementary
\setcounter{table}{0}
\setcounter{figure}{0}
\setcounter{page}{1}
\renewcommand{\thetable}{A\arabic{table}}
\renewcommand{\thefigure}{A\arabic{figure}}
\definecolor{customblue}{rgb}{0.25, 0.41, 0.88} %
{
\setcounter{tocdepth}{2}   %
\hypersetup{linkcolor=customblue}
\tableofcontents
}
\section{Dataset Details}
Below we report the data statistics of our \dataname dataset, detail the data preprocessing steps, pipeline configurations used for each dataset in our experiments, and additional data pipeline experiments that utilize 3D instance mask predictions in caption generation process.

\subsection{Data Statistics}
In Tab.~\ref{tab:datasets}, we report the statistics of our generated dataset, including the number of scenes, RGB-D frames, generated captions, and total tokens in captions for each source dataset. Our dataset contains over 30K scenes, and 5.6M captions with a total of 30M tokens across both real and synthetic indoor environments.
\begin{table}[h]
    \centering
    \resizebox{\linewidth}{!}{
        \begin{tabular}{lrrrrc}
            \toprule
            Dataset & \# Scenes & \# Frames & \# Captions & \# Tokens & Category \\
            \midrule
            ScanNet~\cite{dai2017scannet} & 1,513 & 2.5M & 1.3M & 7.2M & Real \\
            Matterport3D~\cite{chang2017matterport3d} & 2,194 & 0.2M & 0.7M & 3.8M & Real \\
            ARKitScenes~\cite{baruch2021arkitscenes} & 5,045 & 4.0M & 2.4M & 12.6M & Real \\
            ScanNet++~\cite{yeshwanth2023scannet++} & 380 & 0.2M & 0.2M & 0.9M & Real \\
            Structured3D~\cite{zheng2020structured3d} & 20,065 & 0.2M & 1.0M & 5.4M & Synthetic \\
            \midrule
            Total & 29,197 & 7.1M & 5.6M & 29.9M & \\
            \bottomrule
        \end{tabular}
    }
    \caption{\textbf{Statistics of our generated dataset.}
        We report the number of scenes, RGB-D frames, generated captions, and total tokens in captions for each source dataset.
    }
    \label{tab:datasets}
\end{table}

In Tab.~\ref{tab:existing_dataset}, we evaluate caption and 3D mask quality across datasets using three metrics. 
The \textit{unique normalized nouns count} measures the total number of unique normalized nouns in captions, with higher count indicating richer and more diverse captions.
\textit{Mask coverage} (\%) calculates the mean percentage of 3D points with associated captions per scene, where higher coverage enables more effective training.
\textit{Mask entropy} (bits) measures mask quality for datasets with partial masks generated from multi-view images (\ie OV3D, RegionPLC, and Mosaic3D-5.6M) without using GT.
It calculates Shannon entropy of GT instance ID distributions within each mask--higher entropy indicates that a mask contains multiple GT instances, suggesting less accurate mask boundaries.
Mosaic3D-5.6M demonstrates superior caption diversity and mask quality compared to both existing large-scale 3D-text datasets and previous open-vocabulary 3D segmentation datasets, validating its value as a new dataset.
\begin{table}[t]%
    \centering
    \resizebox{\linewidth}{!}{
        \setlength\tabcolsep{2pt}
        \begin{tabular}{@{}l|c|rrr|c|r|rrr@{}}
            \multirow{2}{*}{Train dataset} & Used & \multicolumn{3}{c|}{Mask-Caption Quality} & ScanNet20 & \multicolumn{4}{c}{ScanNet200} \\
            & ScanNet GT & \# Nouns & Coverage & Entropy & f-mIoU & f-mIoU & Head & Com. & Tail \\
            \midrule
            \multicolumn{7}{l}{\textit{Datasets using only ScanNet as source}} \\
            OV3D & \xmark & 2.5K & 70.6 & 72.8 & 45.6 & 7.0 & 18.6 & 2.1 & 0.1 \\
            RegionPLC & \xmark & 1.4K & 77.3 & 81.0 & 50.4 & 8.5 & 21.1 & 3.6 & 0.7 \\
            \rowcolor{gray!15} Mosaic3D-SN\tablefootnote{A subset of Mosaic3D-5.6M using only ScanNet as source dataset.} & \xmark & \underline{9.0K} &  \underline{92.6} & \textbf{60.7} & 65.0 & 13.0 & 30.2 & 6.9 & \underline{1.4} \\
            \midrule
            \multicolumn{7}{l}{\textit{Datasets using multiple sources}} \\
            LEO & \checkmark & 2.6K & 66.2 & - & 65.9 & \underline{14.8} & \textbf{34.3} & \underline{8.3} & \underline{1.4} \\
            SceneVerse & \checkmark & 8.8K & 60.0 & - & \underline{67.3} & 13.6 & 32.4 & 7.3 & 0.8 \\
            EmbodiedScan & \checkmark & 0.3K & 14.0 & - & 44.8 & 6.7 & 16.1 & 3.6 & 0.2 \\
            MMScan & \checkmark & 6.0K & 48.0 & - & 64.1 & 11.7 & 26.1 & 7.9 & 0.7 \\
            \rowcolor{gray!15} Mosaic3D-5.6M & \xmark & \textbf{29.9K} & \textbf{93.7} & - & \textbf{68.1} & \textbf{15.7} & \underline{32.9} & \textbf{10.8} & \textbf{2.7} \\
        \end{tabular}
    }
    \vspace{-4mm}
    \caption{\textbf{Dataset comparison.} We analyze mask-caption quality metrics and annotation-free 3D semantic segmentation performance of different training datasets, while keeping the same model architecture (SpUNet-34C), CLIP model (Recap-CLIP), and loss function (Contrastive).}
    \label{tab:existing_dataset}
    \vspace{-2mm}
\end{table}

\subsection{Data Preprocessing}
\begin{itemize}[leftmargin=*,itemsep=1pt]
    \item \textbf{ScanNet}~\cite{dai2017scannet} To optimize computational efficiency while maintaining adequate spatial coverage, we process every 20th RGB-D frame from each scene. Prior to processing, we resize all RGB-D frames to 640$\times$480 resolution.
    \item \textbf{ScanNet++}~\cite{yeshwanth2023scannet++} From the official dataset, we utilize the \textit{``DSLR''} image collection. Following repository guidelines, we generate synthetic depth images using the reconstructed mesh and camera parameters. After correcting for distortion in both RGB and depth images and adjusting camera intrinsics, we process every 10th frame through our annotation pipeline. Point clouds are generated via surface sampling on the reconstructed meshes.
    \item \textbf{ARKitScenes}~\cite{baruch2021arkitscenes} We leverage the \textit{``3D Object Detection (3DOD)''} subset, utilizing its RGB-D frames and reconstructed meshes. We use every 10th frame at low resolution (256$\times$192), and apply surface point sampling on mesh for point clouds.
    \item \textbf{Matterport3D}~\cite{chang2017matterport3d} We use preprocesed RGB-D frames and point clouds provided by the author of OpenScene~\cite{Peng2023OpenScene}.
    \item \textbf{Structured3D}~\cite{zheng2020structured3d} We utilize RGB-D frames from both perspective and panoramic camera. We utilize preprocessed point clouds from the \textit{Pointcept}~\cite{pointcept2023} library, which fuses multi-view depth unprojection with voxel downsampling to get point clouds.
\end{itemize}

\subsection{Pipeline Configurations}
Our data generation pipeline leverages multiple Visual Foundation Models to automate the data annotation process. Below we detail the configuration of each model in our pipeline.
\begin{itemize}[leftmargin=*,itemsep=1pt]
    \item \textbf{RAM++~\cite{ram_pp}}: we utilize the official pretrained checkpoint \texttt{ram\_plus\_swin\_large\_14m} available at \url{https://huggingface.co/xinyu1205/recognize-anything-plus-model}.
    \item \textbf{Grounded-SAM~\cite{ren2024grounded}}: We employ the official checkpoint of Grounding-DINO~\cite{liu2023grounding} \texttt{IDEA-Research/grounding-dino-tiny} accessed through HuggingFace at \url{https://huggingface.co/IDEA-Research/grounding-dino-tiny}, together with SAM2~\cite{ravi2024sam} with checkpoint \texttt{sam2\_hiera\_l}, available at \url{https://huggingface.co/facebook/sam2-hiera-large}. For the postprocessing, we process the output bounding boxes from Grounding-DINO using a box score threshold of 0.25 and a text score threshold of 0.2. We then apply non-maximum suppression (NMS) with an IoU threshold of 0.5 to remove redundancy. To ensure meaningful region proposals, we filter out excessively large boxes that occupy more than 95\% of the image area. These refined bounding boxes are then passed to SAM2 for mask prediction.
    \item \textbf{Osprey~\cite{yuan2024osprey}}: We utilize the official pretrained \texttt{sunshine-lwt/Osprey-Chat-7b} checkpoint, available at \url{https://huggingface.co/sunshine-lwt/Osprey-Chat-7b}. The generation parameters are set with a temperature of 1.0, top\_p of 1.0, beam search size of 1, and the maximum number of new tokens to 512.
\end{itemize}

\begin{table}[ht]
    \centering
    \begin{minipage}{0.99\columnwidth}\vspace{0mm}    \centering
    \begin{tcolorbox} 
        \raggedright
        \small
        $\texttt{\textbf{System}}$: \texttt{A chat between a curious human and an artificial intelligence assistant. The assistant gives helpful, detailed, and polite answers to the human's questions.} \\
        $\texttt{\textbf{User}}$: \PredSty{\texttt{<image>}} \texttt{This provides an overview of the picture. Please give me a short description of} \PredSty{\texttt{<mask><pos>}} \texttt{, using a short phrase.}
    \end{tcolorbox}
        \label{tab:osprey_prompt}
    \end{minipage}
    \caption{\textbf{Osprey region caption prompt}. Osprey~\cite{yuan2024osprey} utilizes this prompt along with segmentation masks generated by Grounded-SAM to produce descriptive captions for each region.}
    \vspace{-4mm}
\end{table}

\subsection{Additional Pipeline Experiments}
We explore two additional data pipeline configurations that use Segment3D~\cite{huang2024segment3d} masks for segmentation while maintaining Osprey~\cite{yuan2024osprey} for captioning:
\begin{itemize}[leftmargin=*,itemsep=1pt]
    \item \textbf{Segment3D}: We utilize complete Segment3D masks and obtain captions by aggregating descriptions from multiple projected views of each mask. This approach maintains mask completeness but may result in multiple captions being assigned to a single mask from different viewpoints.
    \item \textbf{Segment3D - Mosaic}: We use partial Segment3D masks as seen from individual views and generate captions based on these view-specific projections. While masks are partial, each mask-caption pair is aligned since it represents the exact visible region from a specific viewpoint.
\end{itemize}
The results in Tab.~\ref{tab:segment3d_pipeline} demonstrate that Segment3D - Mosaic outperforms the baseline Segment3D approach, highlighting the importance of precise mask-text pair alignment.
However, both Segment3D variants are outperformed by our \nickname pipeline, which suggests that our combination of RAM++~\cite{ram_pp}, Grounded-SAM~\cite{ren2024grounded}, and SEEM~\cite{zou2024segment} provides superior segmentation quality.

\begin{table}[h]
\centering
\resizebox{\linewidth}{!}{
\begin{tabular}{l|rr|rr}
\midrule
\multirow{2}{*}{Pipeline} & \multicolumn{2}{c|}{ScanNet20~\cite{dai2017scannet}} & \multicolumn{2}{c}{ScanNet200~\cite{scannet200}} \\
 & f-mIoU & f-mAcc & f-mIoU & f-mAcc \\
\midrule
Segment3D~\cite{huang2024segment3d} & 50.6 & 76.6 & 8.3 & 19.1 \\
Segment3D~\cite{huang2024segment3d} - Mosaic & 57.3 & 79.6 & 10.6 & 22.8 \\
\rowcolor{gray!15} \nickname & \textbf{65.0} & \textbf{82.5} & \textbf{13.0} & \textbf{24.5} \\
\midrule
\end{tabular}
}
\caption{\textbf{Segment3D data pipeline evaluation results.}}
\label{tab:segment3d_pipeline}
\end{table}

\section{OV3D~\cite{jiang2024open} Implementation Details}

Since there is no publicly available code and data for OV3D~\cite{jiang2024open}, we utilized our re-implemented version of OV3D for data visualization (Fig.~\ref{fig:data_comparison}) and statistics (Fig.~\ref{fig:data_stats}) in the main manuscript.
In this section, we provide detailed explanations of our re-implementation results.

\subsection{Caption Generation}
OV3D~\cite{jiang2024open} obtains entity-level text descriptions of an image through multi-round conversations with LLaVA-1.5~\cite{liu2024improved}:

\begin{enumerate}
    \item In the first round, LLaVA-1.5 is prompted to generate an image caption describing the overall scene.
    \item In the second round, LLaVA-1.5 is prompted to extract entity names from the generated image caption.
    \item In the final round, LLaVA-1.5 is prompted to generate detailed entity descriptions for each extracted entity name.
\end{enumerate}
During our implementation, we encountered inconsistencies in LLaVA-1.5's response formats. 
To ensure structured and consistent entity-level text descriptions, we modified the final prompt to request responses in JSON format, as shown in Table~\ref{tab:ov3d_prompt}, while maintaining the original prompts for the first two rounds.

\begin{table}[h]
    \centering
    \begin{minipage}{0.99\columnwidth}
        \vspace{0mm}
        \centering
        \begin{tcolorbox} 
            \raggedright
            \small
            $\texttt{\textbf{User}}$: \texttt{Please describe each of the above things that appear in the image with three different nouns or phrases.
            Format your response as a JSON object with the object names as keys and the list of three nouns or phrases as values.
            For example: \{"entity name A": ["description A1", "description A2", "description A3"], "entity name B": ["description B1", "description B2", "description B3"],..\}} \\
            $\texttt{\textbf{Assistant}}$: \texttt{Here is the dictionary of the concrete objects and background classes in the image:}
        \end{tcolorbox}
    \end{minipage}
    \caption{
        \textbf{Modified OV3D entity description prompt.}
        We modified the original OV3D~\cite{jiang2024open} prompt to request JSON responses for consistent entity descriptions.
        For brevity, we omit the previous conversation history that is included in the actual prompt.
    }
    \label{tab:ov3d_prompt}
    \vspace{-4mm}
\end{table}

In addition, our experimental results in Table~\ref{tab:ov3d_reimpl} revealed that LLaVA-1.5's performance in entity name detection was suboptimal, which significantly impacts OV3D's overall effectiveness.
To overcome this limitation, we introduce OV3D++, an enhanced version that uses RAM++~\cite{ram_pp}'s robust tagging capabilities for entity detection while preserving the original entity description process, as shown in Table~\ref{tab:ov3dpp_prompt}.

\begin{table}[h]
    \centering
    \begin{minipage}{0.99\columnwidth}
        \vspace{0mm}
        \centering
        \begin{tcolorbox} 
            \raggedright
            \small
            $\texttt{\textbf{User}}$: \texttt{This is a list of entities, including concrete objects and background classes, in the image: \PredSty{\texttt{<tag>}}.
            Based on your description and the given list of entities, please describe each entity with three different nouns or phrases.
            Format your response as a JSON object with the object names as keys and the list of three nouns or phrases as values.
            For example: \{"entity name A": ["description A1", "description A2", "description A3"], "entity name B": ["description B1", "description B2", "description B3"],..\}} \\
            $\texttt{\textbf{Assistant}}$: \texttt{Here is the dictionary of the concrete objects and background classes in the image:}
        \end{tcolorbox}
    \end{minipage}
    \caption{
        \textbf{OV3D++ entity description prompt with tags.}
        We use RAM++~\cite{ram_pp}'s image tagging output results as the placeholder <tag> to leverage its robust entity detection capabilities.
        For brevity, we omit the previous conversation history that is included in the actual prompt.
    }
    \label{tab:ov3dpp_prompt}
    \vspace{-4mm}
\end{table}

\subsection{Training Objectives}
We experiment with three different training objectives to reproduce OV3D~\cite{jiang2024open}'s performance:

\begin{itemize}
    \item \texttt{DenseAlign}: The original dense alignment loss proposed in OV3D, which maximizes the similarity between text embeddings and point-wise visual features.
    \item \texttt{Align}: A simplified version of dense alignment that computes similarity between text embeddings and pooled visual features within the mask region.
    \item \texttt{Contrastive}: A contrastive learning objective proposed in RegionPLC~\cite{yang2024regionplc} that pulls matching text-visual pairs closer while pushing non-matching pairs apart in the embedding space.
\end{itemize}
For fair comparison, we use SparseUNet34C~\cite{mink} as the backbone network architecture across all experiments, which is the same architecture used in \nickname, and maintain identical training configurations with the only variations being in the training objectives and data generation pipelines.

\subsection{Results}
As shown in Table~\ref{tab:ov3d_reimpl}, our direct re-implementation (OV3D-rep) is unable to fully reproduce the performance reported in the original OV3D paper~\cite{jiang2024open}.
However, our improved version (OV3D++) with RAM++~\cite{ram_pp} tagging achieves better results than the original paper in most metrics when using \texttt{Contrastive} loss, except for f-mIoU on ScanNet20~\cite{dai2017scannet}.
Notably, \texttt{Contrastive} loss consistently outperforms other loss functions across all settings, which motivates our choice to use \texttt{Contrastive} loss in \nickname as well.
While OV3D++ shows significant improvements over the baseline, it is ultimately surpassed by \nickname, demonstrating the effectiveness of \nickname data engine in generating more fine-grained and comprehensive captions.

\begin{table}[h]
    \centering
        \resizebox{\linewidth}{!}{
        \begin{tabular}{ll|rr|rr}
            \toprule
            \multirow{2}{*}{Method} & \multirow{2}{*}{Loss} & \multicolumn{2}{c|}{ScanNet20~\cite{dai2017scannet}} & \multicolumn{2}{c}{ScanNet200~\cite{scannet200}} \\
            & & f-mIoU & f-mAcc & f-mIoU & f-mAcc \\
            \midrule
            OV3D~\cite{jiang2024open} & \texttt{DenseAlign} & 64.0 & 76.3 & 8.7 & - \\
            \midrule
            OV3D-rep & \texttt{DenseAlign} & 34.7 & 54.9 & 4.6 & 8.3 \\
            OV3D-rep & \texttt{Align} & 20.0 & 34.0 & 2.4 & 5.4 \\
            OV3D-rep & \texttt{Contrastive} & 45.6 & 69.8 & 6.9 & 14.3 \\
            \midrule
            OV3D++ & \texttt{DenseAlign} & 54.3 & 71.6 & 7.0 & 12.0 \\
            OV3D++ & \texttt{Align} & 22.5 & 37.6 & 3.1 & 5.6 \\
            OV3D++ & \texttt{Contrastive} & 58.4 & 76.7 & 9.2 & 16.7 \\
            \midrule
            \rowcolor{gray!15}\nickname & \texttt{Contrastive} & \textbf{65.0} & \textbf{82.5} & \textbf{13.0} & \textbf{24.5} \\
            \bottomrule
        \end{tabular}
        }
        \caption{
            \textbf{Re-implementation and improvement of OV3D~\cite{jiang2024open}.}
            We present our re-implementation results of OV3D with three different training objectives: \texttt{DenseAlign}, \texttt{Align}, and \texttt{Contrastive}.
            OV3D-rep denotes our re-implementation, while OV3D++ is our improved version using RAM++~\cite{ram_pp} tagging.
        }
    \label{tab:ov3d_reimpl}
\end{table}

\section{Experimental Analysis}

\subsection{Model Scaling }
\begin{figure}[h]
    \centering
    \includegraphics[width=0.95\linewidth]{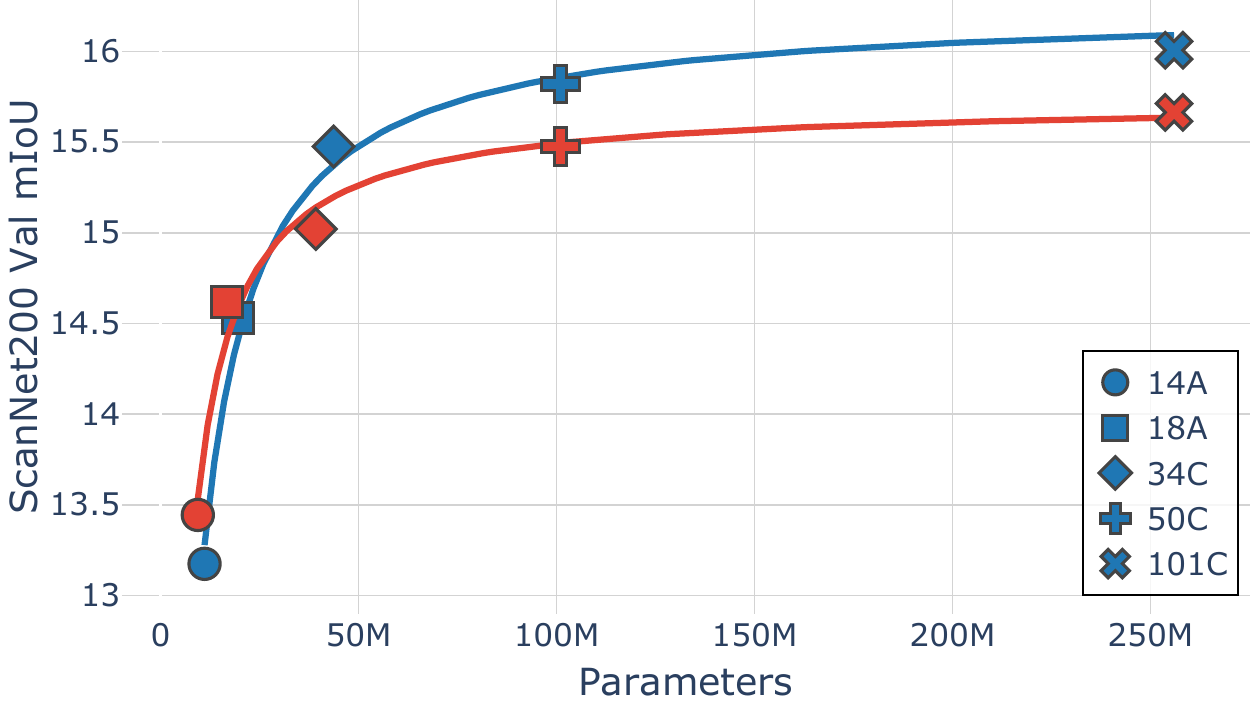}
    \vspace{-2mm}
    \caption{
        \textbf{Model performance scales with model size.}
        We observe consistent improvements in open-vocabulary semantic segmentation on ScanNet200~\cite{scannet200} as we increase the amount of training data.
        This shows the value of our large-scale data generation pipeline in improving open-vocabulary 3D scene understanding.
    }
    \label{fig:model_scaling}
\end{figure}

\noindentbold{Model capacity}
Building on the data scaling analysis, we additionally examine how model scales impact performance. 
We systematically increase the model sizes of 3D encoders while keeping other components fixed. 
We vary the size of Sparse ConvUNet by changing the model depth and widths following literature~\cite{he2016deep}, where the smallest model, SPUNet14A, has 11.1M trainable parameters, whereas the largest variants, SPUNet101C, has 256M parameters.
For these experiments, we fix the training dataset to include ScanNet, ARKitScenes, and ScanNet++.
As shown in Fig.~\ref{fig:model_scaling}, increasing model capacity generally leads to better performance, with diminishing returns after 100M parameters.

\noindentbold{Multi-dataset synergistic learning with PPT~\cite{wu2023towards}}
Since our \nickname dataset combines multiple datasets with different capture settings and environments, there potentially exists domain gaps between each subset that could hinder effective joint training.
Recent work by Wu~\etal~\cite{wu2023towards} demonstrates that adapting dataset-specific learnable prompts in normalization layers can reduce negative transfer effects when training on multiple point cloud datasets.
Building on this insight, we adopt their Point Prompt Training (PPT) approach to ehance our joint training process.
As shown in Fig~\ref{fig:model_scaling}, models using PPT demonstrate better scaling compared to standard joint training, confirming PPT's effectiveness in harmonizing multi-source training on our dataset.

\subsection{Impact of Text Encoders}
To analyze how different text encoders affect open-vocabulary 3D segmentation performance, we evaluate various CLIP text encoders while keeping the 3D encoder architecture (SPUNet34C) and other components fixed. 
Table~\ref{tab:clip_variants} presents the zero-shot performance on ScanNet20 and ScanNet200 benchmarks. 
We compare standard CLIP text encoders including CLIP/B32, CLIP/B16, and CLIP/L14@336px~\cite{radfordLearningTransferableVisual2021}, as well as recently proposed variants like Recap-CLIP~\cite{li2024if} and SigLIP~\cite{zhai2023sigmoid}.

Among all variants, Recap-CLIP achieves the best overall performance with 68.1\% f-mIoU on ScanNet20 and 15.7\% f-mIoU on ScanNet200. 
This represents a +0.3\% and +0.9\% improvement over the base CLIP/B16 model respectively. 
The superior performance of Recap-CLIP aligns with its enhanced text-image alignment ability demonstrated in 2D vision tasks.
Based on these comprehensive experiments, we select Recap-CLIP as our default text encoder for all subsequent experiments. To ensure fair comparisons with previous work, we maintain consistency by using the same text encoder configuration when reproducing baseline results, as shown in Tables~\ref{tab:scannet200_semantic},~\ref{tab:pipeline_comparison}, and~\ref{tab:ov3d_reimpl}. This standardization enables direct performance comparisons and validates the improvements achieved by our proposed approach.

\begin{table}[h]
\centering
\resizebox{\linewidth}{!}{
\begin{tabular}{l|cc|cc}
\midrule
\multirow{2}{*}{CLIP Model} & \multicolumn{2}{c|}{ScanNet20~\cite{dai2017scannet}} & \multicolumn{2}{c}{ScanNet200~\cite{scannet200}} \\
 & f-mIoU & f-mAcc & f-mIoU & f-mAcc \\
\midrule
CLIP/B16~\cite{radfordLearningTransferableVisual2021} & 67.1 & 83.8 & 14.4 & 27.7 \\
CLIP/B32~\cite{radfordLearningTransferableVisual2021} & \underline{67.8} & \underline{84.5} & 14.8 & 26.5 \\
CLIP/L14@336px~\cite{radfordLearningTransferableVisual2021} & 64.2 & 81.9 & 14.9 & 27.7 \\
SigLIP~\cite{zhai2023sigmoid} & 66.3 & \textbf{84.6} & \underline{15.3} & \textbf{29.0} \\
Recap-CLIP~\cite{li2024if} & \textbf{68.1} & 84.4 & \textbf{15.7} & \underline{28.3} \\
\midrule
\end{tabular}
}
\caption{\textbf{Impact of CLIP text encoders on open-vocabulary 3D semantic segmentation.} We train our SPUNet34C architecture on the full \dataname dataset (5 subsets) with different CLIP text encoders while keeping other components fixed. Recap-CLIP~\cite{li2024if} achieves the best overall performance across both ScanNet20 and ScanNet200 benchmarks, demonstrating the importance of text encoder selection for zero-shot generalization.}
\label{tab:clip_variants}
\end{table}

\subsection{Annotation-free 3D Referring Segmentation}
To quantitatively analyze the attention between free-form text queries and point features shown in Fig.~\ref{fig:qual}, we leverage the 3D referring segmentation annotations from ScanRefer~\cite{chen2020scanrefer}. 
This allows us to evaluate how well our model's attention aligns with human-annotated referring expressions in 3D scenes.
Specifically, we evaluate our model's zero-shot performance on the ScanRefer validation set without any fine-tuning on the 3D referring segmentation task.
For each referring expression in ScanRefer, we use it as a text query to obtain attention maps between the query and point features.
We then threshold the cosine similarity scores to obtain binary segmentation masks, where points with positive similarity scores (greater than 0) are considered as the predicted region.
The predicted masks are compared against ground truth annotations using standard IoU metrics.

As shown in Table~\ref{tab:scanrefer}, our method outperforms both OpenScene-3D~\cite{Peng2023OpenScene} and RegionPLC~\cite{yang2024regionplc}, demonstrating its superior ability to highlight relevant regions for free-form text queries.
These results demonstrate that our model not only excels at semantic segmentation with simple class names but also achieves superior zero-shot performance on more complex free-form referring expressions, quantitatively validating its effectiveness as a general-purpose 3D vision-language foundation model.

\begin{table}[h]
    \centering
    \resizebox{\linewidth}{!}{
        \begin{tabular}{lccc}
            \toprule
            Method & OpenScene-3D$^{\dagger}$~\cite{Peng2023OpenScene} & RegionPLC$^{\flat}$~\cite{yang2024regionplc} & \nickname \\
            \midrule
            mIoU & 3.1 & 3.7 & 5.3 \\
            \bottomrule
        \end{tabular}
    }
    \caption{
        \textbf{Annotation-free 3D referring segmentation on ScanRefer~\cite{chen2020scanrefer}.}
        $^{\dagger}$ and $^{\flat}$ denote official checkpoints and our reproductions, respectively.
    }
    \label{tab:scanrefer}
\end{table}

\section{Additional Results}

\subsection{Quantitative Results}
In Tab.~\ref{tab:scannet200_category}, 
We conduct a comprehensive evaluation of our model's performance across different category frequencies in ScanNet200. 
Following standard practice~\cite{scannet200}, we categorize labels into head, common, and tail groups based on their occurrence frequency in the dataset.  
As shown in Tab.~\ref{tab:scannet200_category}, our approach achieves consistent improvements across all category groups compared to previous methods.
Notably, we observe the relative gain is more substantial on common and tail categories as we incorporate more training datasets, highlighting the effectiveness of our multi-dataset training strategy in learning robust features across varying scene distributions.
\begin{table}[h]
    \centering
    \resizebox{\linewidth}{!}{
    \begin{tabular}{l|rrr}
    \midrule
    \multirow{2}{*}{Method} & \multicolumn{3}{c}{ScanNet200 val mIoU (\%)} \\
    & Head & Common & Tail \\
    \midrule
    OpenScene-3D$^{\dagger}$ & 16.4 & 2.6 & 0.2 \\
    RegionPLC$^{\flat}$ & 24.2 & 2.7 & 0.4 \\
    \midrule
     \rowcolor{gray!15}\nickname & & & \\
    - SN~\cite{dai2017scannet} & 30.2 & 6.9 & 1.4 \\
    - SN~\cite{dai2017scannet} + AR~\cite{baruch2021arkitscenes} & 32.4 & 9.3 & 2.0 \\ 
    - SN~\cite{dai2017scannet} + AR~\cite{baruch2021arkitscenes} + SN2~\cite{yeshwanth2023scannet++} & 33.3 & 10.0 & 2.6 \\
    - SN~\cite{dai2017scannet} + AR~\cite{baruch2021arkitscenes} + SN2~\cite{yeshwanth2023scannet++} + M~\cite{chang2017matterport3d} & 32.9 & 10.0 & 2.5 \\
    - SN~\cite{dai2017scannet} + AR~\cite{baruch2021arkitscenes} + SN2~\cite{yeshwanth2023scannet++} + M~\cite{chang2017matterport3d} + S3D~\cite{zheng2020structured3d} & 32.9 & 10.8 & 2.7 \\ 
    \midrule
    \end{tabular}
    }
    \caption{\textbf{Category-wise performance analysis on ScanNet200~\cite{scannet200}}. $^{\dagger}$ and $^{\flat}$ denote official checkpoints and our reproductions, respectively.}
    \label{tab:scannet200_category}
\end{table}

\subsection{Qualitative Results}

\begin{figure*}[htbp]
    \centering
    \begin{minipage}{\textwidth}
        \begin{subfigure}[b]{0.45\textwidth}
            \includegraphics[width=\textwidth,trim={7cm 5cm 9cm 1cm},clip]{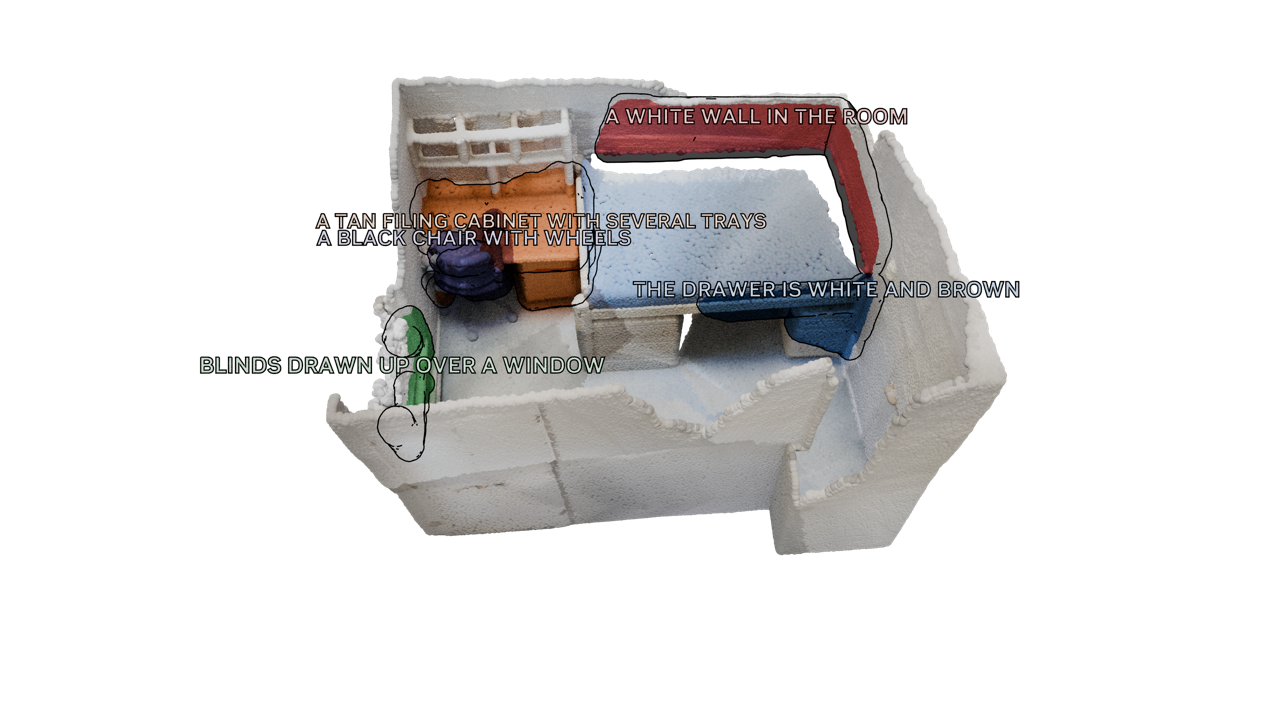}
            \caption{scene0055\_02 (ScanNet)}
            \label{fig:scene0055}
        \end{subfigure}%
        \hfill%
        \begin{subfigure}[b]{0.45\textwidth}
            \includegraphics[width=\textwidth,,trim={8cm 5cm 6cm 3cm},clip]{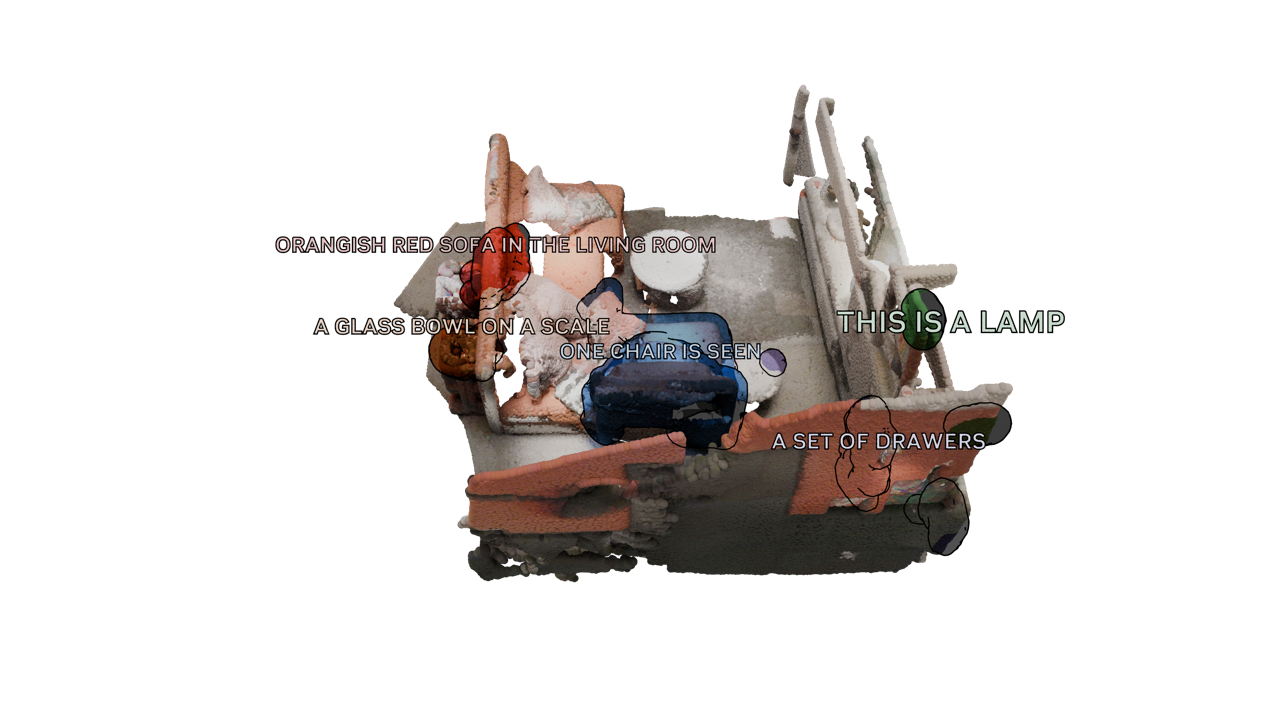}
            \caption{scene0128\_00 (ScanNet)}
            \label{fig:scene0128}
        \end{subfigure}
    \end{minipage}
    
    \vspace{10pt}  %
    
    \begin{minipage}{\textwidth}
        \begin{subfigure}[b]{0.45\textwidth}
            \includegraphics[width=\textwidth,,trim={8cm 1cm 8cm 0cm},clip]{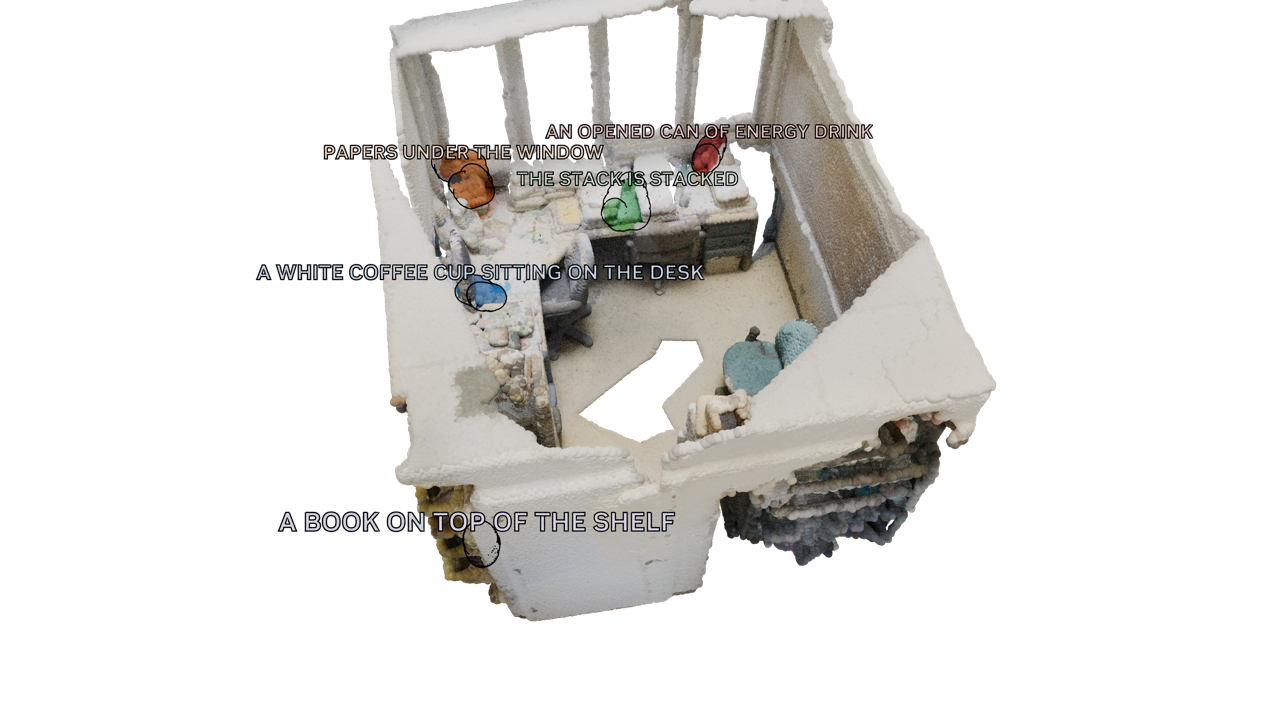}
            \caption{scene0211\_01 (ScanNet)}
            \label{fig:scene0211}
        \end{subfigure}%
        \hfill%
        \begin{subfigure}[b]{0.45\textwidth}
            \includegraphics[width=\textwidth,trim={8cm 3cm 8cm 0cm},clip]{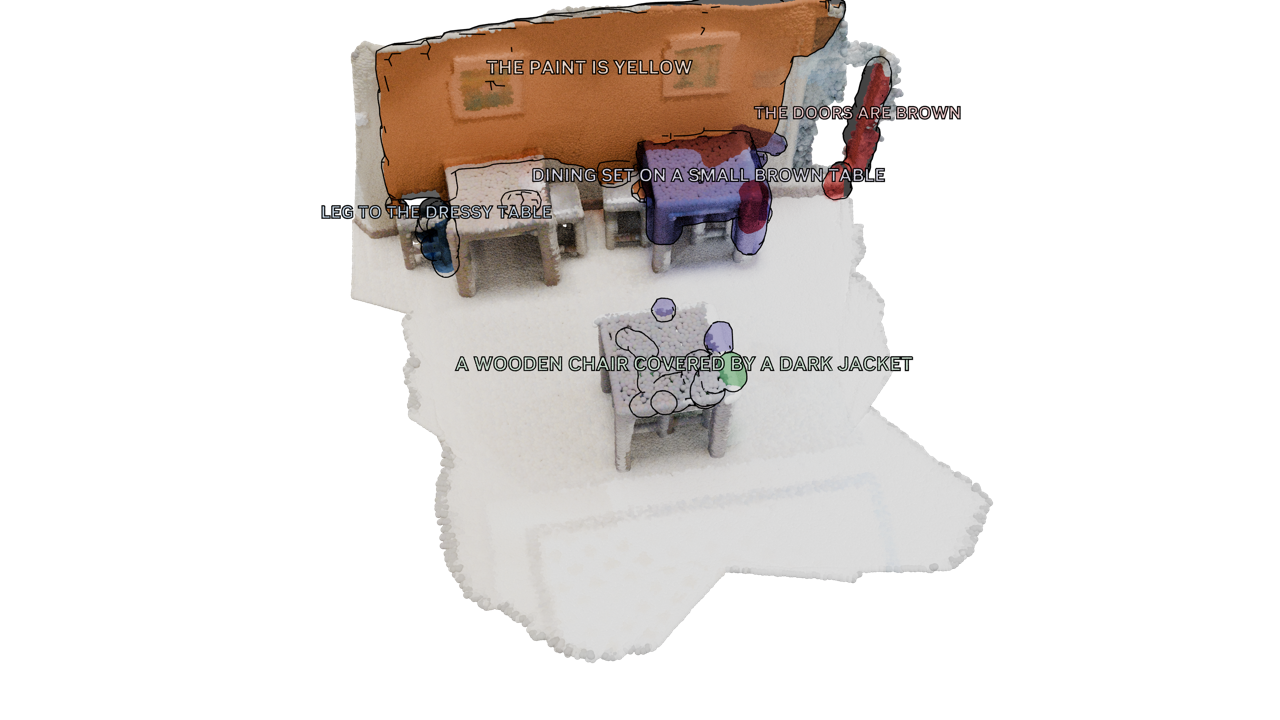}
            \caption{scene0324\_00 (ScanNet)}
            \label{fig:scene0324}
        \end{subfigure}
    \end{minipage}

    \begin{minipage}{\textwidth}
        \begin{subfigure}[b]{0.45\textwidth}
            \includegraphics[width=\textwidth,,trim={8cm 1cm 6cm 0cm},clip]{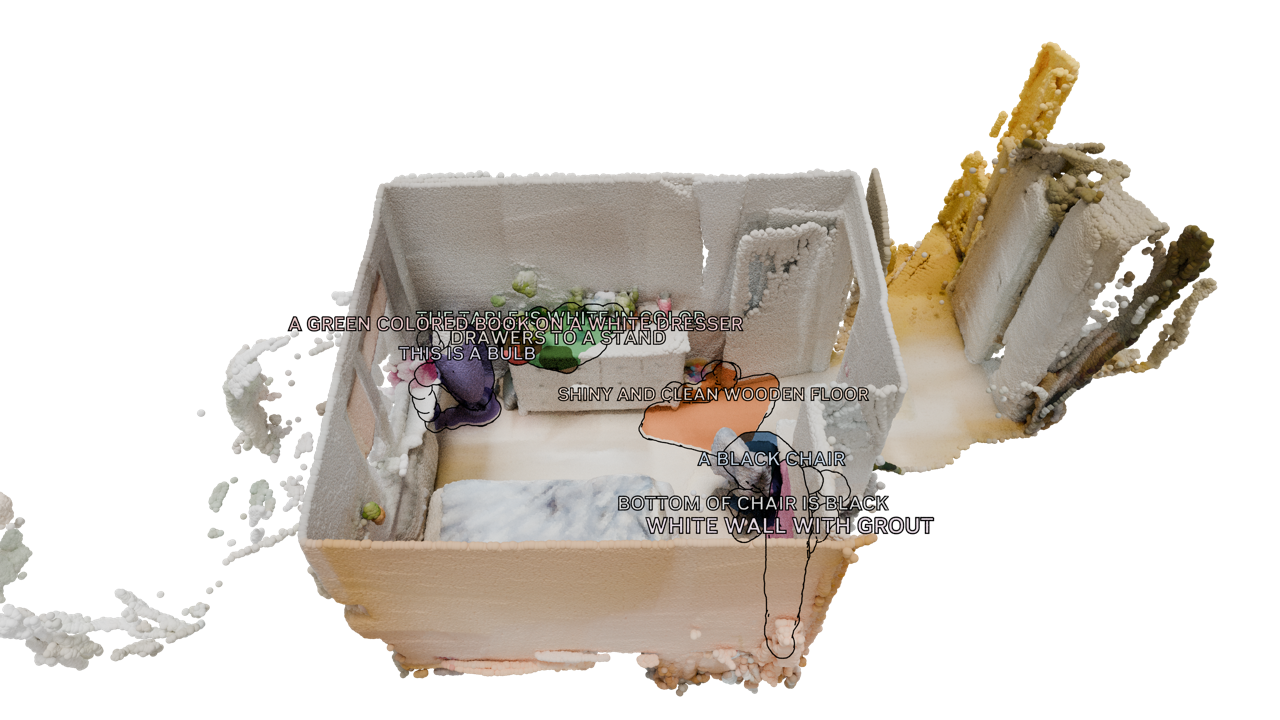}
            \caption{47333055 (ARKitScenes)}
            \label{fig:47333055}
        \end{subfigure}%
        \hfill%
        \begin{subfigure}[b]{0.45\textwidth}
            \includegraphics[width=\textwidth,trim={5cm 1cm 8cm 0cm},clip]{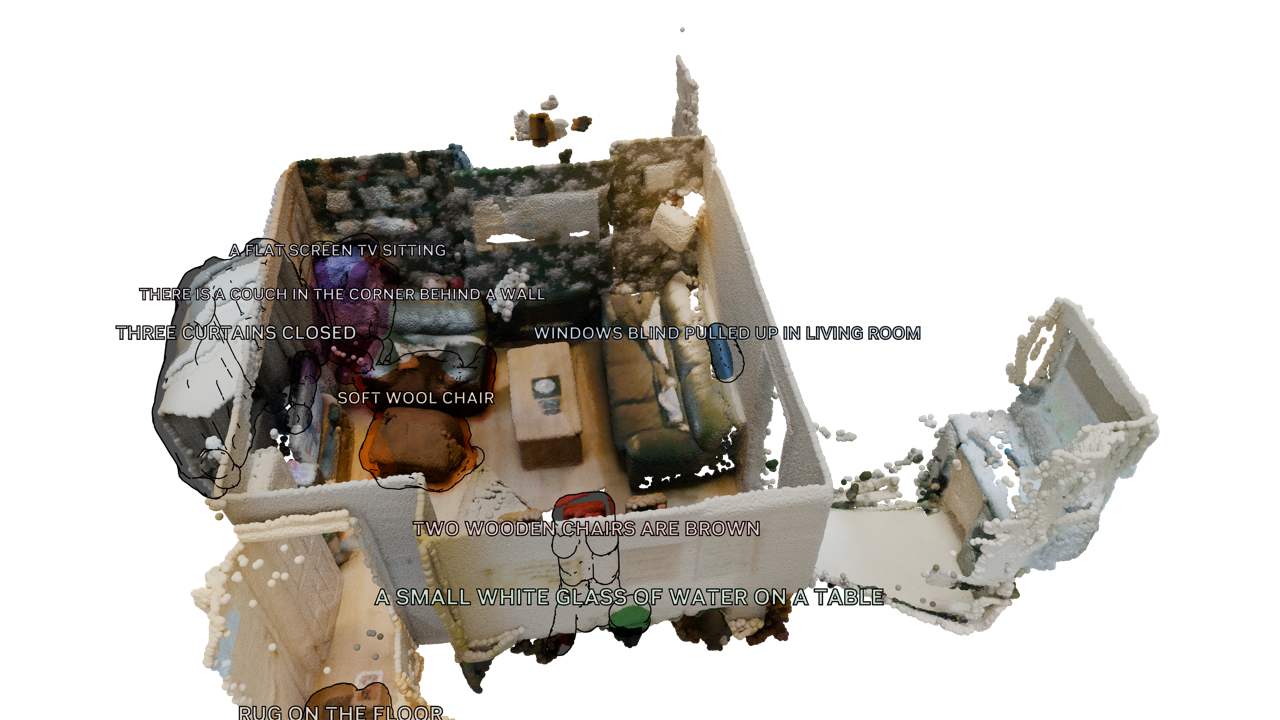}
            \caption{42898477 (ARKitScenes)}
            \label{fig:42899477}
        \end{subfigure}
    \end{minipage}
    
    \caption{\textbf{More visualization of the 3D mask-text pairs in our \dataname dataset.} A subset of mask-text pairs has been chosen for better visualization.}
    \label{fig:caption_viz_all}
\end{figure*}

\begin{figure*}[t!]
    \centering
    \includegraphics[width=\linewidth]{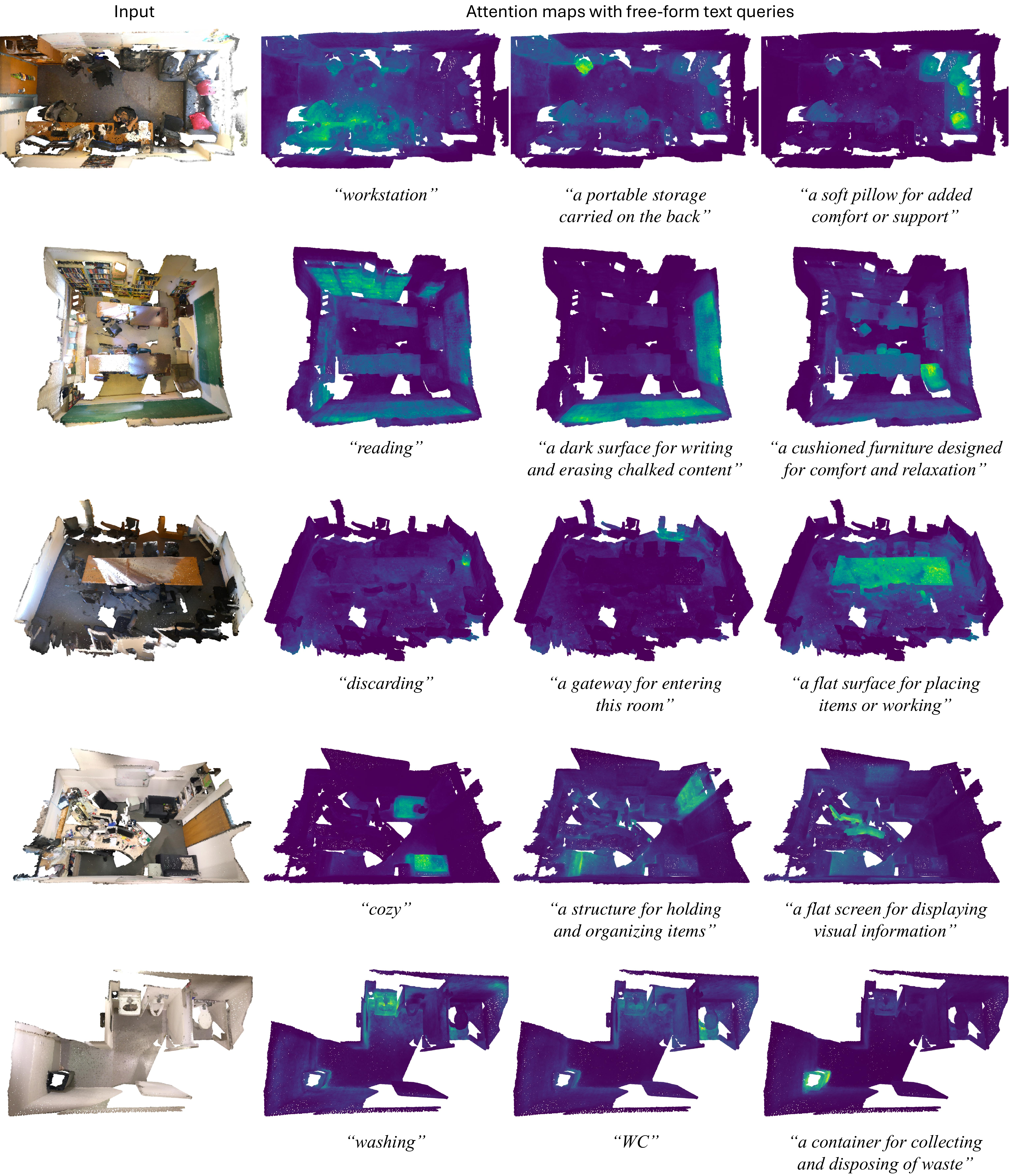}
    \caption{
        \textbf{Attention visualization of \nickname as a 3D foundational model.}
        We observe that our model can highlight relevant regions even without explicitly mentioning ScanNet~\cite{dai2017scannet,scannet200} class names in queries.
        The model also effectively attends to regions related to abstract concepts like affordances (\eg, reading, discarding, washing).
    }
    \label{fig:supp_qual}
\end{figure*}

\begin{figure*}[t!]
    \centering
    \includegraphics[width=\linewidth]{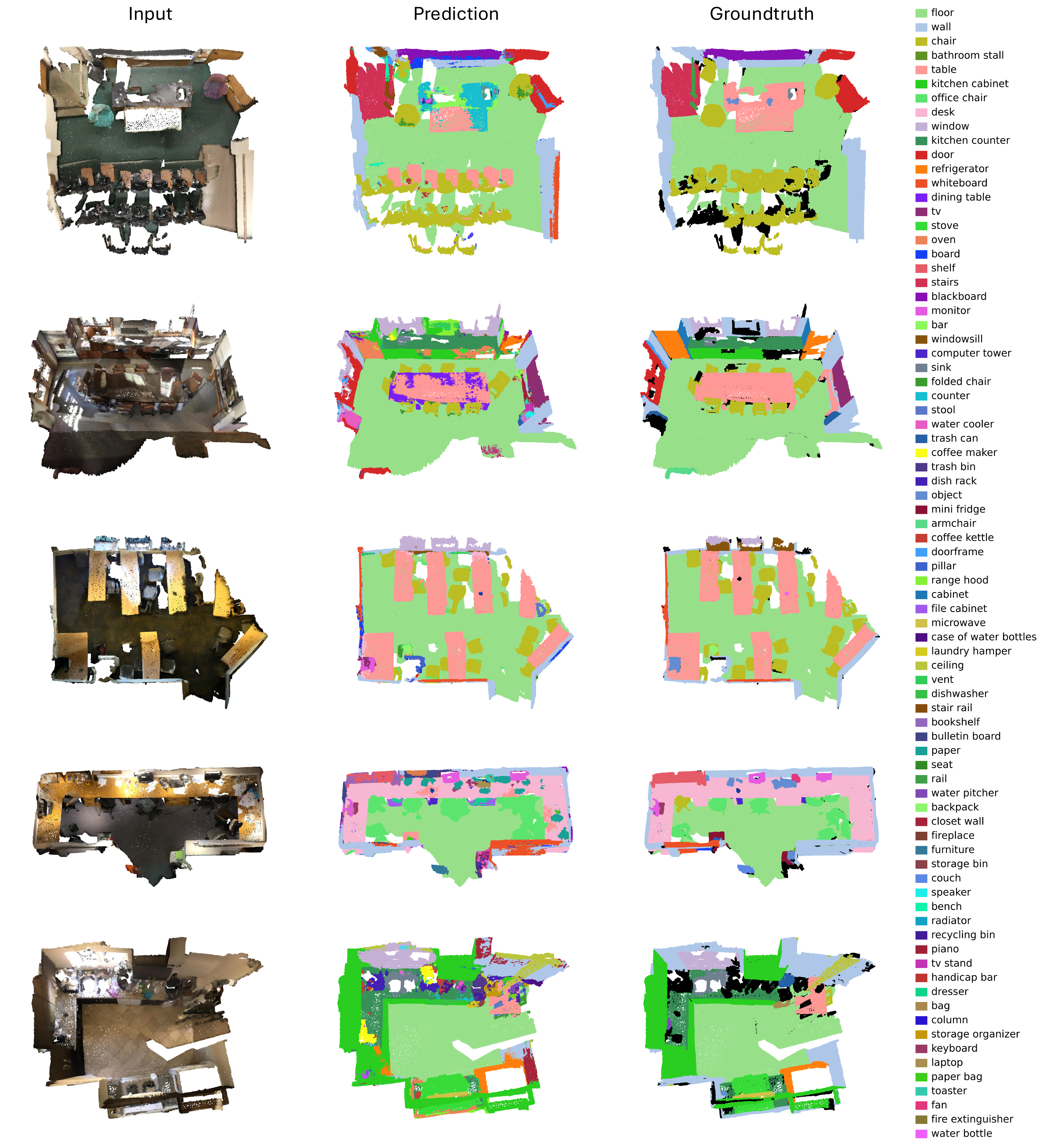}
    \caption{
        \textbf{Qualitative results of annotation-free 3D semantic segmentation on ScanNet200~\cite{scannet200}.}
        Despite being trained without ground truth annotations, \nickname shows competitive results on ScanNet200~\cite{scannet200}.
    }
    \label{fig:supp_qual_sem}
\end{figure*}

In Fig.~\ref{fig:caption_viz_all}, we present additional qualitative visualizations of our generated 3D mask-text pair datasets, where we carefully selected mask-text pairs to effectively demonstrate the diversity and quality of our generated data.
Furthermore, in Fig.~\ref{fig:supp_qual}, we showcase attention maps for diverse text queries across various scenes, which demonstrates that our model can effectively attend to relevant regions in response to different types of queries, ranging from object-centric descriptions to more abstract concepts like affordances.
In Fig.~\ref{fig:supp_qual_sem}, we present qualitative results of annotation-free 3D semantic segmentation on ScanNet200~\cite{scannet200}.
Our model shows promising results, particularly in the first scene where it demonstrates an interesting behavior - while the ground truth annotates an integrated chair-desk unit entirely as a chair, \nickname distinctly separates and predicts the desk and chair components.
This showcases a potential advantage of our annotation-free approach to training 3D foundation models, where the model can learn more nuanced semantic distinctions that might be overlooked in manual annotations.

\clearpage

\clearpage

\clearpage

\clearpage

\end{document}